%% file: TNNLS.tex
\documentclass[journal]{IEEEtran}
\IEEEoverridecommandlockouts
\usepackage[noadjust]{cite}
\usepackage{amsmath,amssymb,amsfonts}
\usepackage{algorithmic}
\usepackage{graphicx}
\usepackage{textcomp}
\usepackage[table]{xcolor} % Rudi: added [table] for row colours
\def\BibTeX{{\rm B\kern-.05em{\sc i\kern-.025em b}\kern-.08em
    T\kern-.1667em\lower.7ex\hbox{E}\kern-.125emX}}

\usepackage{hyperref}
\usepackage{float}
\usepackage[sort, capitalise]{cleveref} % options: compress and noabbrev

\usepackage{graphics}
\usepackage{multirow}
\usepackage{tikz}
\usepackage[inline]{enumitem}

\ifCLASSOPTIONcompsoc
    \usepackage[caption=false, font=normalsize, labelfont=sf, textfont=sf]{subfig}
\else
\usepackage[caption=false, font=footnotesize]{subfig}
\fi

\newcommand*\circled[1]{\tikz[baseline=(char.base)]{
    \node[shape=circle,draw,inner sep=1pt] (char) {#1};}}
    
\usepackage{balance}

\begin{document}

\title{\LARGE \bf
Identifying Appropriate Intellectual Property Protection Mechanisms\\
for Machine Learning Models: A Systematization of\\
Watermarking, Fingerprinting, Model Access, and Attacks 
}

\author{Isabell Lederer$^{1}$ and Rudolf Mayer$^{2}$ and Andreas Rauber$^{3}$% <-this % stops a
%space
\thanks{
This work was partially funded by the European Union’s Horizon 2020 research and innovation programme under grant agreement no. 826078 (project 'FeatureCloud').
This publication reflects only the authors’ view and the European Commission is not responsible for any use that may be made of the information it contains.
SBA Research (SBA-K1) is a COMET Center within the COMET – Competence Centers for Excellent Technologies Programme and funded by BMK, BMAW, and the federal state of Vienna. The COMET Programme is managed by FFG.
}
\thanks{$^{1}$Isabell Lederer was with SBA Research, Vienna, Austria while working on this paper.}
\thanks{$^{2}$Rudolf Mayer is with SBA Research, Vienna, Austria, and the Institute of Information Systems Engineering, Faculty of Informatics, Vienna University of Technology, Vienna, Austria. Email:
{rmayer@sba-research.org}}
\thanks{$^{3}$Andreas Rauber is with SBA Research, Vienna, Austria, and the Institute of Information Systems Engineering, Faculty of Informatics, Vienna University of Technology, Vienna, Austria. Email:
{andreas.rauber@tuwien.ac.at}}
}
\maketitle

\begin{abstract}
The commercial use of Machine Learning (ML) is spreading; at the same time, ML models are becoming more complex and more expensive to train, which makes Intellectual Property Protection (IPP) of trained models a pressing issue. Unlike other domains that can build on a solid understanding of the threats, attacks and defenses available to protect their IP, the ML-related research in this regard is still very fragmented. This is also due to a missing unified view as well as a common taxonomy of these aspects.

In this paper, we systematize our findings on IPP in ML, while focusing on threats and attacks identified and defenses proposed at the time of writing.
We develop a comprehensive threat model for IP in ML, categorizing attacks and defenses within a unified and consolidated taxonomy, thus bridging research from both the ML and security communities.
\end{abstract}


%%%%%%%%%%%%%%%%%%%%%%%%%%%%%%%%%%%%%%%%%%%%%%%%%%%%%%%%%%%%%%%%%%%%%%%%%%%%

\section{Introduction}

In many Machine Learning (ML) settings, training an effective model from scratch -- especially complex and powerful models such as a Deep Neural Network (DNN) -- (i) is computationally very expensive, (ii) requires expertise for setting parameters, and (iii) the amount of data needed is not commonly accessible or expensive to obtain. Security concerns become more prominent when these models are made available to other parties or customers, e.g., in Machine Learning as a Service (MLaaS) or under a license. This is when model owners -- who have invested significant resources to train a model and now want to offer it to customers -- start to consider Intellectual Property Protection (IPP) methods like watermarking (to verify ownership) and access control (to prevent unauthorized usage of a model).
IP litigation cases over ML models do occur, but have so far not seen widespread media attention; however, protection mechanisms are therefore investigated from a legal point of view, e.g., \cite{michiels_how_2020}, showing that the burden of proof is generally lies with the IP rights holder. Thus, it is important to anticipate the need for such proofs and protect ML models with IPP techniques.

In the last few years, we have consequently seen an increase in research on IPP techniques for ML models. Many black- and white-box watermarking methods have been proposed based on techniques such as backdoor embedding via data poisoning or regularization.
At the same time, several studies have shown the vulnerability of some of these schemes against novel attacks. Similar observations hold true for model access control techniques. However, a comprehensive overview on the field, including a unified nomenclature and taxonomy, is still missing.
Based on a systematic review, this paper provides a survey and systematization of knowledge. 
\\
Our contributions are:
\begin{itemize}
    \item A systematic overview on research related to IPP of ML, focusing on reactive (watermarking and fingerprinting) and proactive (e.g., model access) techniques
    \item A taxonomy to categorize ML IPP schemes
    \item A categorization of 36 approaches by a set of characteristics identified through methodological comparison
    \item An analysis of vulnerability to attacks designed to break the IPP schemes
    \item Guidelines on how to choose a fitting watermarking/IPP scheme for a given setting
    \item A framework for implementations of watermarking methods and available trained and watermarked models, allowing to compare other methods to previous research\footnote{Available at \url{https://sbaresearch.github.io/model-watermarking/}}
\end{itemize}

The remainder of this paper is structured as follows: \Cref{sec:relatedWork} provides an overview of related surveys.
Our research methodology is described in \cref{sec:method}.
\cref{sec:background} provides definitions and background to machine learning, deep neural networks, watermarking, and fingerprinting.
\cref{sec:ippOverview} introduces our taxonomy of IPP methods, the threat model, and attacks.
\cref{sec:watermarking,sec:fingerprinting} then discuss current approaches for watermarking and fingerprinting schemes, while
\cref{sec:otherIPP} discusses proactive IPP methods such as access control.
\cref{sec:attacks} provides a taxonomy of currently known attacks and which IPP methods are vulnerable to them.
In \cref{sec:guidelines} we provide guidelines for choosing fitting IPP methods in various scenarios, before we provide our conclusions in
\cref{sec:conclusions}.

\section{Related Work} \label{sec:relatedWork}
As the first work in this field, Chen et al. \cite{chen_performance_2018} empirically investigate five watermarking schemes for ML models (two white- and three black-box),
evaluate of their fidelity, and estimate the robustness against three attacks (model fine-tuning, parameter pruning, and watermark overwriting), thus providing an important early comparison of these techniques' effectiveness.
We expand on this and provide a survey as well as systematization of the overall IP protection field for ML models. %
Concurrently to our work, a survey specifically covering the watermarking of Machine Learning models was published as a pre-print by Boenisch \cite{boenisch_survey_2020}. Watermarking is an important aspect, which our work complements with fingerprinting and proactive methods such as model access, thus providing a holistic view of the entire IPP field.

\section{Methodology}
\label{sec:method}

\subsection{Literature Search} \label{appendix:method:search}

\begin{figure} %h!
 \centering
 \includegraphics[width = 0.48 \textwidth]{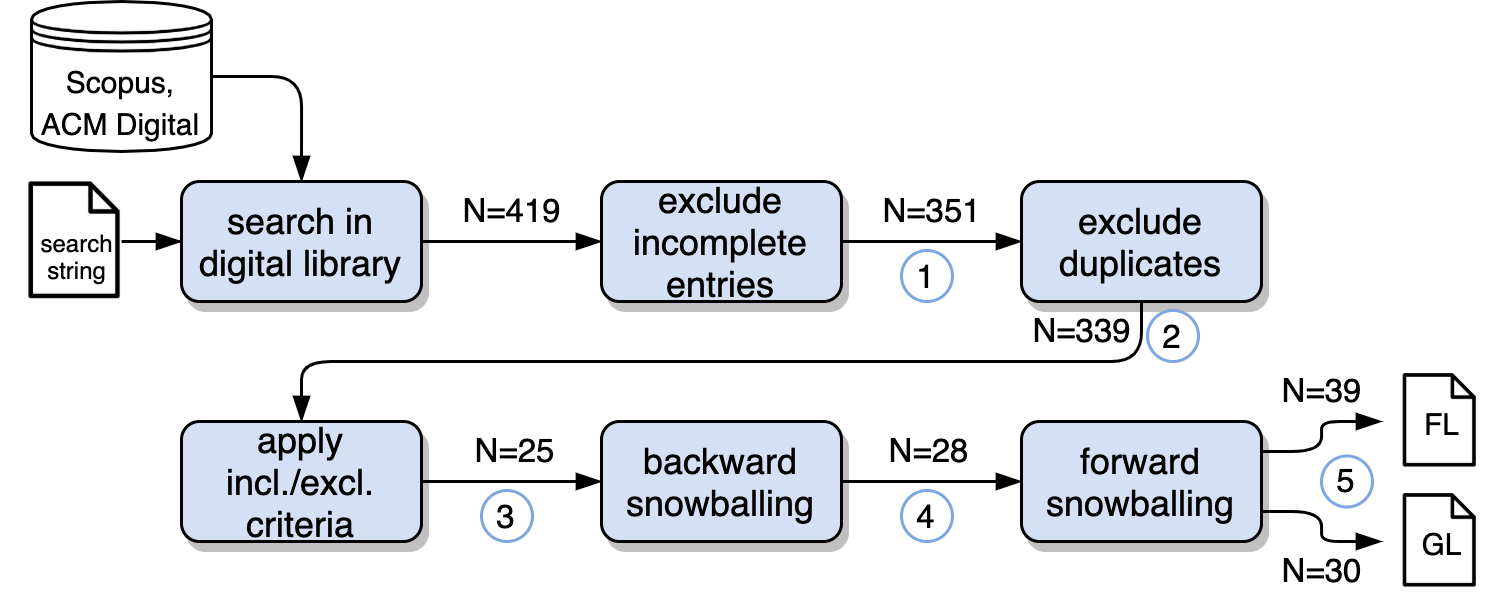}
 \caption{Literature search process workflow. In every step we denote the number of publications by $N=x$. The numbers 1 to 6 correspond to the CSV-files which contain all the retrieved literature in the particular step.}
 \label{fig:search-process}
\end{figure}

In preparation for this systematization we performed an extensive literature search following the guidelines by Kitchenham et al. \cite{kitchenham_guidelines_2007}.
\cref{fig:search-process} shows of our literature search process. 
We distinguish between the following types of publications: \textit{formal literature} (FL), i.e., peer-reviewed literature such as book sections, conference papers, and journal articles; and \textit{gray literature} (GL), i.e., literature that did not undergo a peer review process, for example pre-prints (published e.g. on arXiv, university repositories, personal websites, etc.)

\begin{figure} %h!
 \centering
 \includegraphics[width = 0.48 \textwidth]{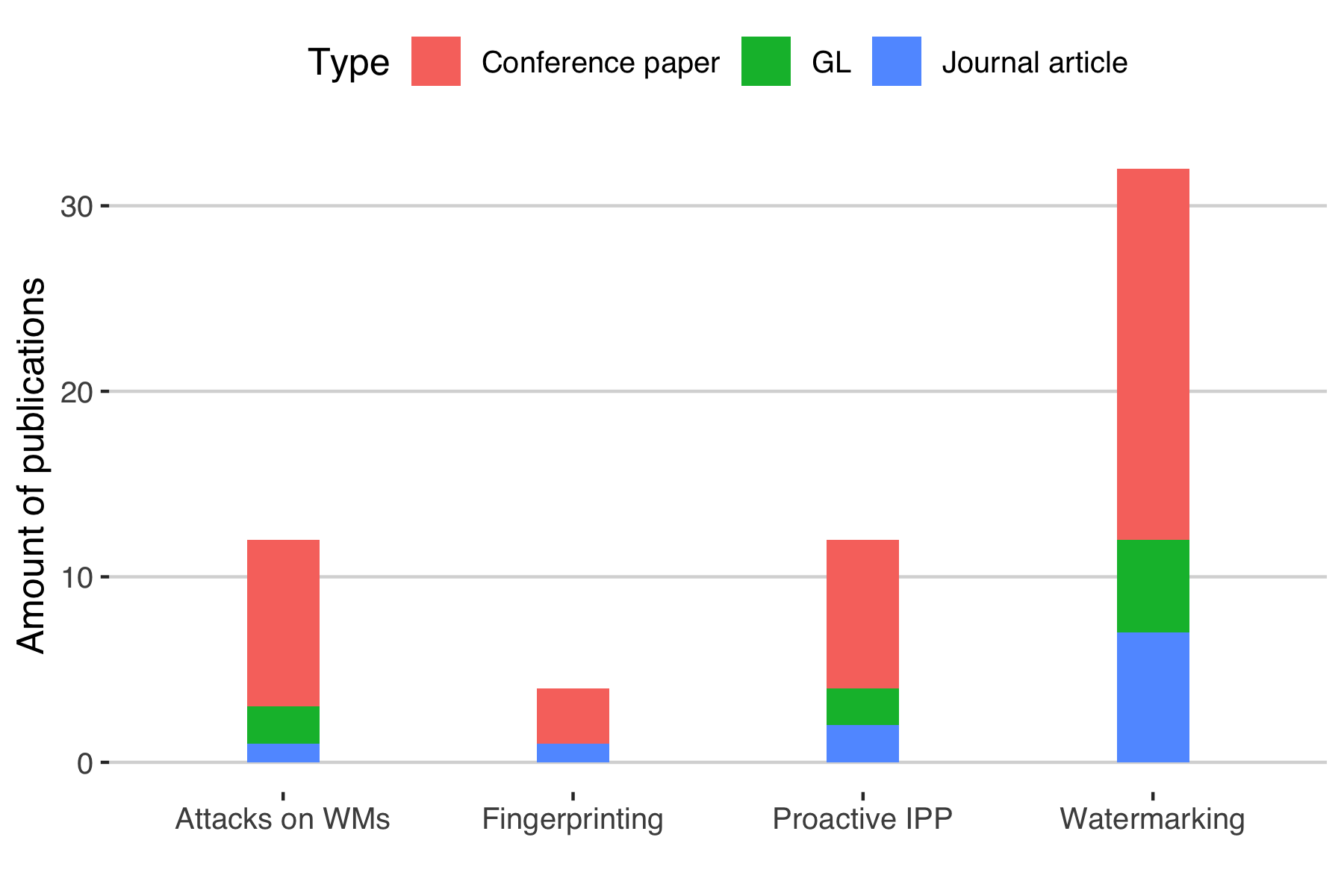} % plt2_transparent.png
 \caption{The literature distribution across different topics regarding IPP of ML models. 
 Most of the papers address watermarking.}
 \label{fig:aufteilung}
\end{figure}

\cref{fig:aufteilung} shows the distribution of publications regarding the different topics and literature types\footnote{The topics will be explained in more detail in \cref{sec:watermarking}--\ref{sec:attacks}}. We can clearly see that most papers address watermarking; however, there is also a significant number of papers on attacks. Note that some publications include both a novel attack to a scheme and a novel watermarking scheme, which is immune to this attack.

\begin{figure} %h!
 \centering
 \includegraphics[width = 0.48 \textwidth]{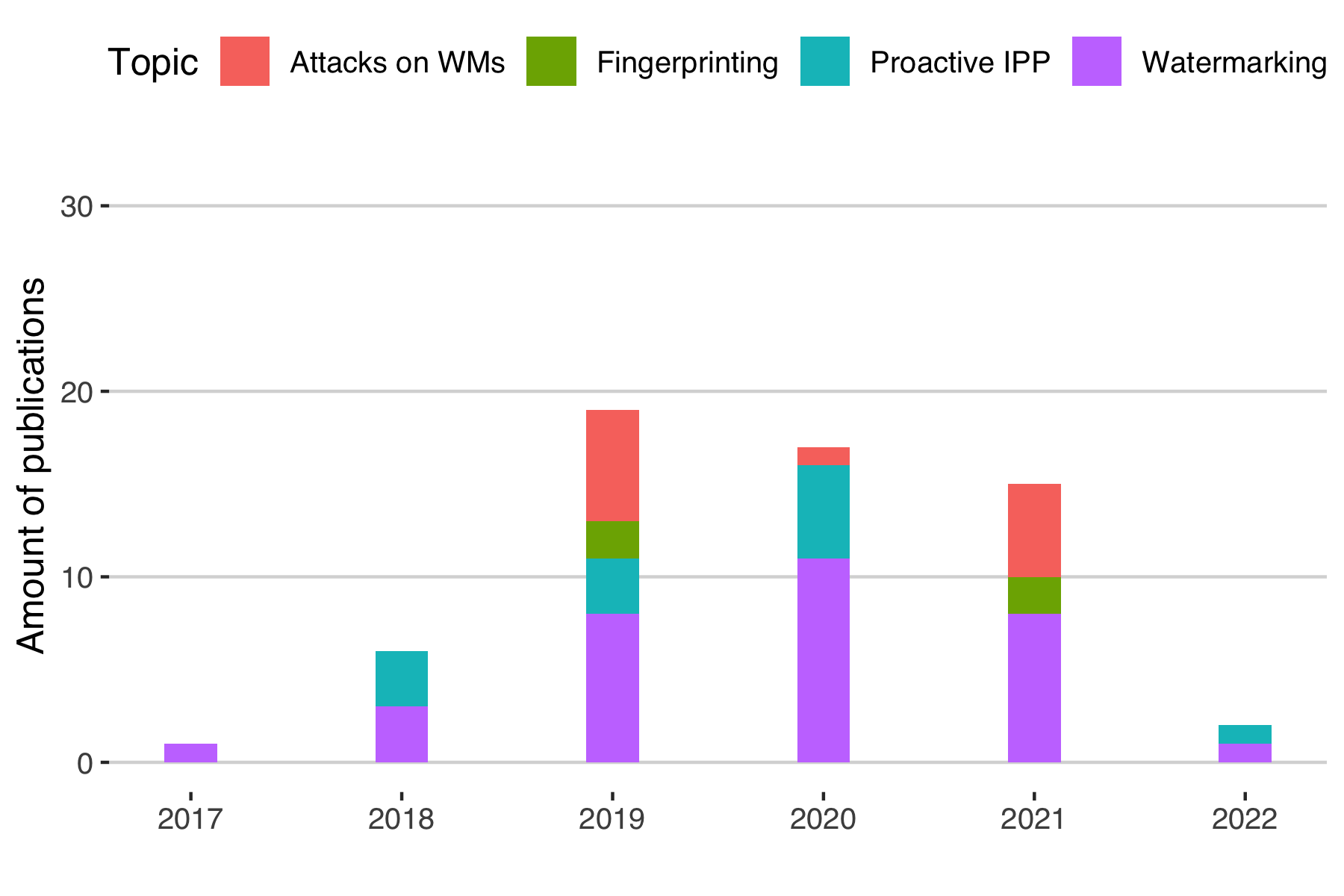} % plt1_transparent.png
 \caption{The literature distribution over the years for different topics.}
 \label{fig:aufteilung-jahre}
\end{figure}

\cref{fig:aufteilung-jahre} shows the distribution of publications across the publishing years. We see a rising research interest for all topics, with papers on attacks being published only recently.

\subsection{Inclusion- and Exclusion Criteria} \label{appendix:method:criteria}

In order to facilitate reproducibility of the literature search, we defined and documented the following criteria to find the most relevant literature covering IPP of ML models. Our inclusion criteria are:
\begin{enumerate*}[label=(\roman*)]
%\begin{itemize}
    \item literature which proposes an IPP scheme for ML models,
    \item literature which proposes an attack on an IPP scheme for ML models, and
    \item literature which evaluates or compares earlier schemes.
%\end{itemize}
\end{enumerate*}

Our exclusion criteria are:
\begin{enumerate*}[label=(\roman*)]
%\begin{itemize}
    \item (near) Duplicates \footnote{If the titles are different, but the content is very similar, we include all versions of this item and indicate that fact. However, we subsequently cite only the most complete version, as suggested by Kitchenham et al. \cite{kitchenham_guidelines_2007}.},
    \item literature which only \textit{uses} ML for multimedia watermarking, such as image watermarking, and
    \item literature that only \textit{applies} previously published IP protection schemes, without a novel or large-scale evaluation.
%\end{itemize}
\end{enumerate*}

\section{Preliminaries} \label{sec:background}

This section provides the necessary background for the remainder of the paper.
In this work, we focus on \textit{supervised} learning, an area of ML including \textit{classification} and \textit{regression}.

Some learning algorithms -- such as the (stochastic) gradient descent commonly employed in Deep Neural Networks (DNNs) or Convolutional Neural Networks (CNNs) -- iteratively adapt their learnable parameters by minimizing some form of loss function.
To prevent overfitting to the training data, a \textit{parameter regularizer} is oftentimes used. This is an additional term in the loss function, often in the form of a penalty that controls the magnitude of the parameter values. 
In the field of model IPP, regularizers are frequently used to embed a watermark into a model.

The process of \textit{Fine-Tuning}, i.e., further training a model on different training data, usually with a smaller learning rate, can be used for either improving the model or, when using it for a slightly different purpose, in \textit{Transfer Learning}. In the context of model IPP, we use it either to embed a watermark or as a malicious modification to a well-trained model to remove unwanted information (e.g., a watermark).

\textit{Knowledge Distillation} \cite{hinton_distilling_2014} is a compression technique that uses knowledge of a model (teacher network) to train a new, smaller, computationally cheaper model (student network).

\textit{Generative Adversarial Networks} (GANs) use two models: one (the generator) learns the actual data generation task, and the other (the discriminator) evaluates it  \cite{goodfellow_generative_2014}.

\textit{Federated Learning} \cite{yang_federated_2019} is a ML technique in which multiple parties are involved to train the model on their data (without sharing the data, mostly to preserve privacy). 

An \textit{Autoencoder} (AE) is a special artificial neural network that is commonly used for dimensionality reduction \cite{hinton_reducing_2006}.
This is achieved by learning to replicate its input to its output via a smaller hidden layer that learns to represent the input.

\textit{Adversarial Examples} \cite{szegedy_intriguing_2014} are inputs created to fool a model. An original input is perturbed by some specially crafted noise such that the model is unable to classify the generated input instance correctly. The perturbation is kept minimal in order to be less noticeable by humans or technical detection methods.

We understand ML-based \textit{Image Processing} as applying a model to an input image, with the output being an image as well.
The model is trained to perform image enhancing, embed unrecognizable data or other transformations such as neural style transfer \cite{jing_neural_2020}. It is important to differentiate image processing from \textit{image pre-processing}, which is usually performed on an image dataset before training a model and includes techniques like resizing, cropping, or normalization.

\subsection{Watermarking}
Digital watermarking is a well-studied method e.g. in multimedia  \cite{kahng_watermarking_1998} and relational databases \cite{kamran_comprehensive_2018} IPP. The main idea is to embed a piece of imperceptible%
\footnote{Perceptible watermarks are also commonly used in the multimedia domain, e.g. logos or copyright notices superimposed on images or videos. Imperceptible watermarks, however, aim to avoid changing the perceptible impression of a work. This is the type of watermark we consider for IPP of ML models.} 
signature in the data (e.g., image or audio) to deter malicious usage.
Digital watermarking is thus a form of steganography or information hiding, i.e., the practice of concealing a message within another message.
The hidden information must be embedded in such a way that no algorithm can remove or overwrite the watermark. More recent digital watermarking techniques (e.g., for images) make use of deep learning techniques in the embedding process \cite{zhong_automated_2020}; similarly, also attacks targeted to remove such watermarks are increasingly using deep learning \cite{sharma_robust_2020}.
Quiring et al. \cite{quiring_adversarial_2018}, for instance, combine methods from model stealing to generate a \textit{substitute} model of a watermark detector, and then generate adversarial examples against this model in order to obtain images with minimal perturbations, thus evading detection.

The watermarking methods we consider in this paper are (ML) model watermarking, i.e., the IP that has to be protected is \textit{the ML model} itself. Model watermarking is related to multimedia watermarking, but the techniques differ since the asset to be protected differs. 
Research on watermarking ML models predominantly addresses image classification (cf. \cref{sec:watermarking}), and thus CNNs. The introduced terminology is thus strongly influenced by this application of ML, but the concepts are transferable to other input types.

\subsection{Fingerprinting}
We consider fingerprinting as an extension of watermarking. While watermarking has the purpose to verify the \textit{owner} of a digital asset, fingerprinting wants to trace its (potentially malicious) \textit{recipient}. Therefore, fingerprinting techniques should be capable of embedding multiple, but unique marks to identify the recipient.
Similar to watermarking, fingerprinting is already widely used in multimedia areas like images, audio, video \cite{lach_fpga_1998}, or digital data stored in relational databases \cite{yingjiu_li_fingerprinting_2005}.

\section{Taxonomy of IPP for ML Models}
\label{sec:ippOverview}

\input{fig-IPP-overview.tex}

In this section, we define our threat model and provide a comprehensive taxonomy of IPP methods for ML models to mitigate the risks posed by those threats.
Subsequently, we give an overview of attacks against those IPP mechanisms. %TODO rephrase

\subsection{Threat Model}
\label{sec:threatmodel}
We first need to understand the motives of an attacker (also called adversary or malicious user).
The entity that invested resources to obtain a ML model for a specific task (``model owner'') wants to offer this model to a certain target audience/customer.
The most prominent reasons for an attacker to (illegally) re-distribute a model are (i) no/not enough training data, expertise, time or computational power to train such a model themselves, and/or (ii) the unwillingness to agree with the license terms of the obtained model or the fees for using Machine Learning as a Service (MLaaS).
We call the model to be protected the \textit{target model} and the attacker's model -- which stems from the target model -- the \textit{adversary model}.
In our threat model, we assume one of the following scenarios:
\begin{enumerate}
    \item \textbf{Legal copy:} The model owner distributes the model publicly, either for free, e.g., via a platform such as Model Zoo \cite{modelzoo}, but with a restrictive license, or for a fee. The attacker re-distributes it, e.g., via a lucrative API service.
    
    \item \textbf{Illegal copy:} The model owner distributes the model as a pay-per-query API service. The attacker performs a Model Stealing (or extraction) attack \cite{tramer_stealing_2016,oliynyk_i_2023} and re-distributes it as above, e.g., via their own API service.
    
%TODO: keep here?
%A specific attack against the IP of ML models is the so-called \textit{model stealing} (or -extraction) attack \cite{oliynyk_i_2023} which aims to reveal a model's internal characteristic or copy a complete model by only querying its API service. The intended target may be the learned parameters, decision boundary, or functionality, as well as the model architecture or other training hyper-parameters.

\end{enumerate}

Regardless of how the attacker obtained the model, in both cases, the IP of the model owner is illegally utilized. 
However, it is important to differentiate between those cases, as this has a large impact on the selection of potential defense mechanisms.

\subsection{IPP Methods}
We developed a comprehensive taxonomy of IPP methods for ML models, depicted in \cref{fig:overview}.
We distinguish between (i) reactive methods which respond to a threat event, and (ii) proactive methods, meaning the defender takes the initiative to prevent a threat event.
Methods that enable to verify the ownership of a model through \textit{model watermarking} and \textit{model fingerprinting} are thus \textit{reactive}; methods that, e.g., seek to prevent unauthorized model access are \textit{proactive}. 
Ownership verification is a weak form of protection, as it requires the unauthorized usage of the model to be known (or at least suspected) and needs some form of access to the model. Model access control, on the other hand, may prevent such illegal use by rendering the model useless for unauthorized users. This is comparable to preventing unauthorized use of, e.g., software.

Some of the methods we introduce in this section can be distinguished by whether they are \textit{white-box} or \textit{black-box}.
White-box means that the model owner needs access to the parameters or other characteristics of the adversary model during the IPP method process, e.g., watermark extraction (cf. \Cref{fig:bothworkflows}).
As this scenario is often unrealistic, black-box mechanisms tend to be more popular.
These generally only need access to the model's prediction -- e.g., via an API service -- to observe matching input and output from the ML model and use it in a similar fashion to an \textit{oracle}.

%TODO: maybe can be skipped? Maybe also moved to some place further down?
Watermarking as defence against Model Stealing attacks (scenario 1) in our threat model) is mostly achieved through specific black-box watermarking techniques which survive such an attack, i.e., the hidden information is "stolen" with the model. In the case that a user legally obtained a copy of the ML model (scenario 2)), but then is using it not according to the licensing terms, more techniques are available. White-box approaches for this case embed the ownership information directly into the model parameters or their probability density function (PDF). Black-box approaches mostly rely on specific input samples, so-called "trigger sets", that will cause the model to behave in a way that is unexpected for the task, and unpredictable to the attacker. 
These techniques mainly differ in how the respective triggers are constructed.

Model access control methods can be distinguished via the asset they want to protect. Most work focuses on the protection of the model parameters, either through encryption, other obfuscation techniques, or by requiring a specific method to transform the inputs. If the model structure (or architecture) is to be protected, usually obfuscation techniques are employed.

\begin{figure*}[t]
\centering
    \input{fig-InformationHiding_ML_Process_wide}
    \caption{Different notions of information hiding along an ML process}
    \label{fig:watermarking-ml-process}
\end{figure*}

Watermarking and fingerprinting of ML models are forms of steganography (information-hiding); however, we want to point out that there are several other connotations for watermarking, and information hiding in general, along the machine learning process (as depicted in \cref{fig:watermarking-ml-process}).
For example, Sablayrolles et al. \cite{sablayrolles_radioactive_2020} propose a technique that \textit{traces data usage}; it marks (training) data so that an ML model trained on that data will bear a watermark that can be identified (cf. \circled{1} in \cref{fig:watermarking-ml-process}).
However, the main body of work regarding watermarking -- and also the respective focus in this paper -- considers ML models as the asset to be protected through embedded watermarks (cf. \circled{2} in \cref{fig:watermarking-ml-process}).
Abdelnabi et al. \cite{abdelnabi_adversarial_2021} are not watermarking a \textit{model}, but the \textit{output} of a (text-)generating model (cf. \circled{3} in \cref{fig:watermarking-ml-process}). They assume that an attacker could use the model to generate entire articles; subsequently, the watermark can be extracted from the generated text and prove an illegitimate use of the model.
For some settings it is further considered that a marked output (prediction or data) is generated with the explicit goal to trace the usage of this data, e.g., by an attacker (cf. \circled{4} in \cref{fig:watermarking-ml-process}). This is a special form of \circled{1}, given that the data origin is different, and of \circled{2}, as the adversary model is \textit{implicitly marked} (cf. %. Most methods following this approach are described in 
\cref{sec:wm:countering_extraction}).

Other forms of steganography may be employed in an attack against machine learning processes. For instance, Song et al. \cite{song_machine_2017} propose a technique to exfiltrate data from a private training dataset by hiding them within the parameters of a model that was trained on this dataset. 
This way, adversaries who do not have direct access to the training data, but are allowed by the data owner to run an ML training algorithm on the data, can exfiltrate this data via the derived machine learning model, i.e., perform a data exfiltration attack (cf. \circled{5} in \cref{fig:watermarking-ml-process}).

In this work, however, we focus on techniques which hide and rightfully ingrain information about a legal owner or -- in case of fingerprinting -- a recipient of the model.

\subsection{Attack Model} \label{sec:attack-model}
This section introduces the attack models, which we will further detail \cref{sec:attacks}. We assume that the attacker obtains a legal copy of the target model, and either knows or suspects that the model has an IPP in place.
We begin with attack models against watermarks, as these are transferable to other IPP types; we consider the following cases:
\\
\textbf{Watermark detection:} The attacker seeks to detect the watermark, potentially to perform a targeted watermark removal or overwriting. If the watermark is not secured with additional mechanism (e.g., a private key for extraction), the attacker could also claim ownership.
\\
\textbf{Watermark overwriting:} The attacker tries to replace the existing watermark with their watermark, thus rendering the model owner's watermark useless.
\\
\textbf{Watermark invalidation:} The attacker aims to disable the watermark function so that it cannot be verified, without actually removing it from the model 
%TODO: remove or keep?
%\footnote{An example is an ensemble attack where API services do not respond relying on a single model's prediction, but on several different models so that when triggering, the expected trigger label is not revealed \cite{hitaj_evasion_2019}.}
\\
\textbf{Watermark removal:} The attacker wants to modify the model in such a way that the model owner's watermark extraction algorithm will no longer result in proving correct ownership. 
\\
Most of these attacks also work against \textit{fingerprinting}.

With regard to \textbf{model access control} mechanisms, detection is often trivial, as an active mechanism will result in low fidelity. An attacker would mostly want to \textbf{remove}, \textbf{invalidate}, or potentially \textit{overwrite} the mechanism to (i) gain unauthorized access to use the model (as black-box), or (ii) to reveal either the model architecture or -parameters for other purposes.

\section{Watermarking of ML models} \label{sec:watermarking}
The vast majority of watermarking methods for ML models is designed specifically for DNNs. The main reason for this is not only the high value of DNNs, as they require large datasets and long training periods, but also the number of "degrees of freedom" in a DNN. Large DNNs thus have, compared to other ML models, more "space" for hiding watermarks. 
Most authors evaluate the schemes for image classification tasks. However, some extend to other tasks, like audio classification \cite{jia_entangled_2021}, image processing \cite{quan_watermarking_2020, wu_watermarking_2021, zhang_model_2020} (the output being an image/data rather than a prediction), or specific settings, such as GANs \cite{yu_artificial_2021}, Federated Learning \cite{tekgul_waffle_2021}, Graph Neural Networks \cite{zhao_watermarking_2021}, and Deep Reinforcement Learning \cite{behzadan_sequential_2019}.

The following terminology is common in model watermarking and used throughout this paper: Watermark \textbf{embedding} is the process of placing the watermark into the model, e.g., via fine-tuning. 
Watermark \textbf{extraction} is the process of extracting the embedded watermark from the model, but neither in a malicious nor permanent way (which are called watermark \textit{detection} and watermark \textit{removal}, respectively). Extraction means to identify if and which watermark has been placed. Subsequently, during watermark \textbf{verification}, the extracted watermark is compared to the model owner's secret to prove ownership. Following certain rules, e.g. thresholding the bit error rate, it is determined if the watermarks are the same.

\begin{figure*}[t]
\centering
\hspace*{\fill}
  \subfloat[\label{fig:whitebox-workflow}]{%
       \includegraphics[width = 0.46 \textwidth]{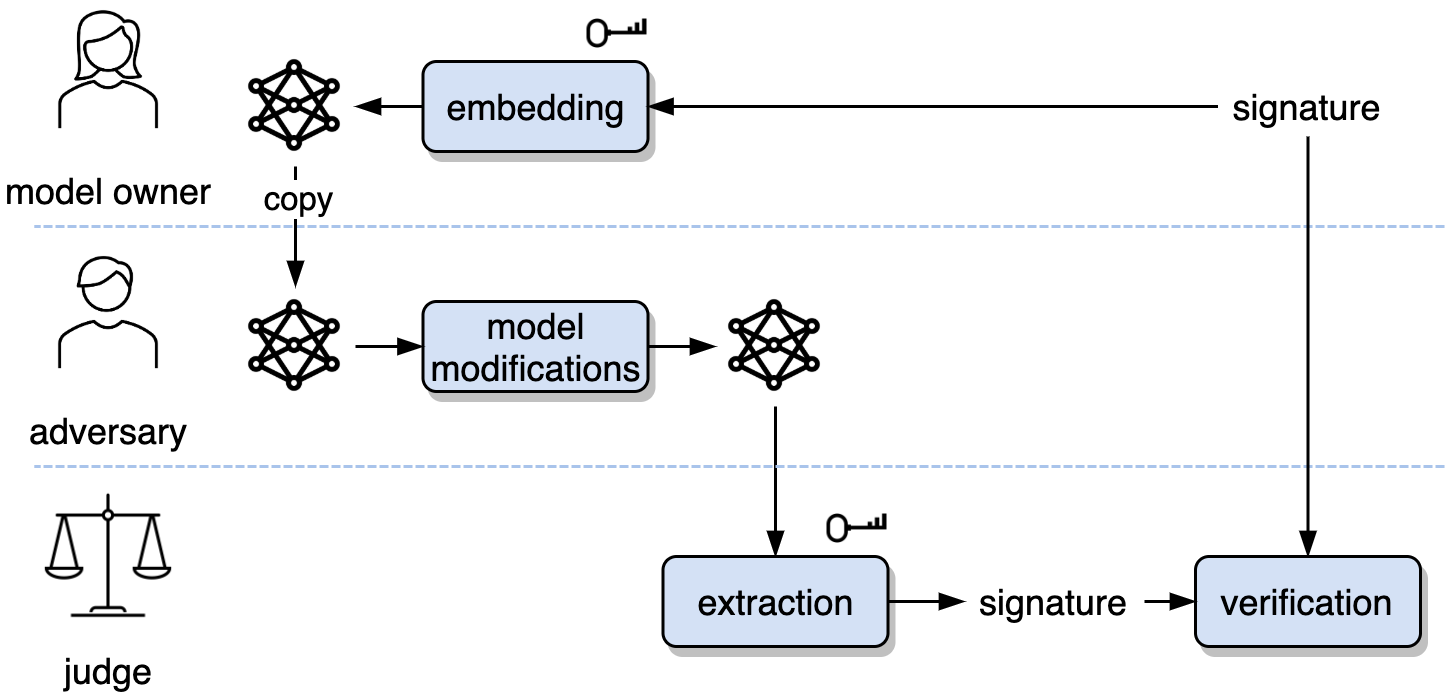}}
    \hfill
  \subfloat[\label{fig:blackbox-workflow}]{%
        \includegraphics[width = 0.46 \textwidth]{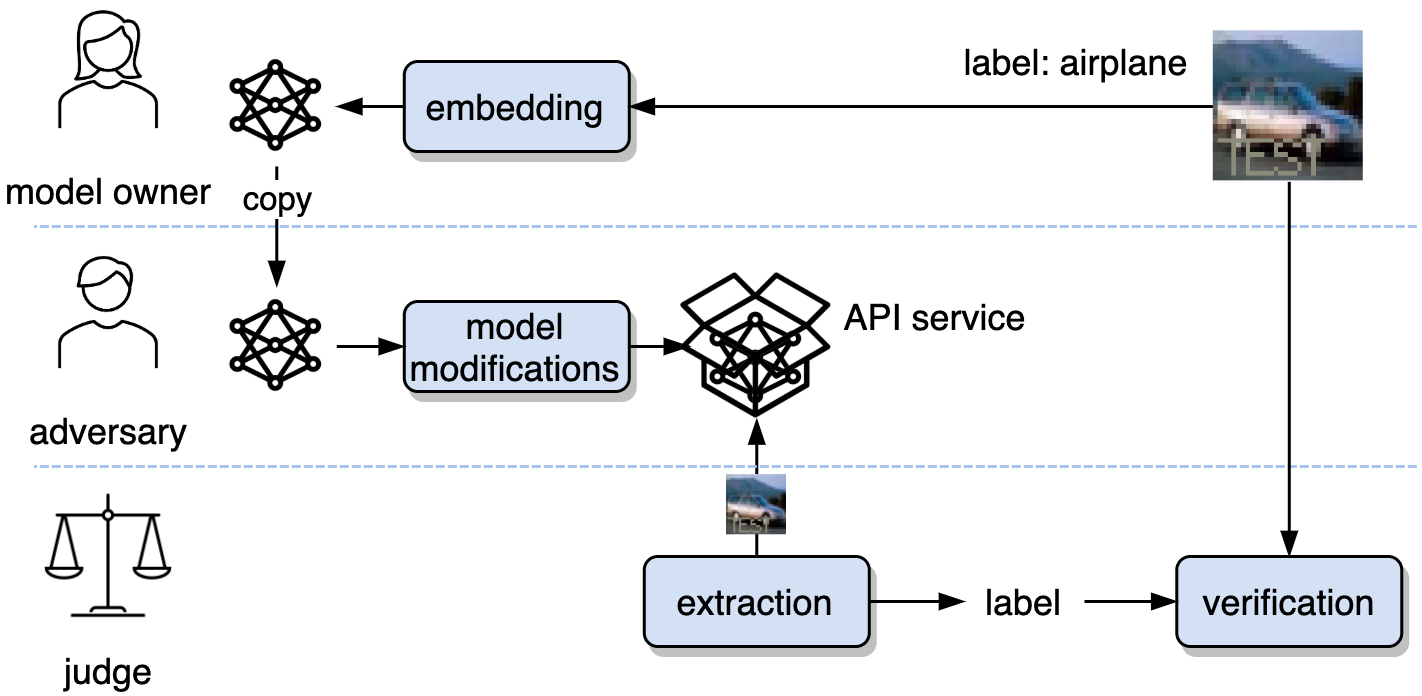}}
        \hspace*{\fill}
    \caption{Typical workflows for (a) white-box watermarking and (b) black-box watermarking}
    \label{fig:bothworkflows}
\end{figure*}

Typical workflows for white-box and black-box watermarking are shown in \cref{fig:whitebox-workflow} and \cref{fig:blackbox-workflow}, respectively.
%
% --- figure both workflows
% ---
%
For white-box watermarking, the model owner creates a $T$-bit signature vector $\mathbf{b} \in \{0,1\}^T$ which is a set of arbitrary binary strings that should be independently and identically distributed (iid) \cite{rouhani_deepsigns_2019}. This binary vector serves as a watermark and is usually embedded into the model through fine-tuning with regularization. We call this type of embedding scheme \textit{regularizer-based} (cf. \cref{sec:whitebox}). Note that other ways of embedding are proposed by Uchida et al. \cite{uchida_embedding_2017}, e.g. during the training or via knowledge distillation.

For black-box watermarking, the model owner creates specially crafted trigger inputs which receive wrong labels on purpose. 
When the model is "triggered" by these inputs, it behaves unexpectedly to a normal user (cf. \cref{sec:blackbox}).

\subsection{Requirements}

Watermarking (and fingerprinting) schemes should fulfill several requirements. Literature is not coherent in terminology; we therefore provide a common nomenclature in this paper. 
To this end, we collected all requirements that are proposed in literature and list them in \cref{tab:requirement}, identifying terms which are used synonymously and referencing respective publications.
\footnote{
Note that there are a few more terms used in literature that can not be easily mapped.
\textit{Feasibility} is a combination of robustness and effectiveness \cite{li_how_2019}, and \textit{correctness} of effectiveness, reliability, and integrity \cite{lukas_deep_2021}. 
\textit{Non-trivial ownership} is used in multiple ways -- sometimes as a synonym for integrity, meaning that innocent models are not being accused of ownership piracy; but also as a requirement that an attacker cannot easily claim ownership without knowing the watermarking scheme and embedded watermark. \textit{Authentication} is rather a subset of effectiveness than a real synonym, since it only requires a provable association between an owner and their watermark.
}

\input{tab-requirements.tex}

The most important requirements are \textbf{effectiveness}: the watermark should be embedded in a way that the model owner can prove ownership anytime; \textbf{fidelity}: the model's accuracy should not be degraded because of the watermark embedding; and \textbf{robustness}: the watermark embedding should be robust against several kinds of attacks, including fine-tuning, model compression, and other attacks specific to certain methods.

Fingerprinting should fulfill two more requirements, namely \textbf{uniqueness}: the fingerprint can be uniquely attributed to a certain user; and \textbf{scalability}: the fingerprinting scheme should be able to embed multiple fingerprints.

\input{tab-summary-req.tex}
We provide an overview of all the watermarking and fingerprinting schemes considered in this paper, and whether they are meeting the above-mentioned requirements, in \cref{tab:summary-req}.
We observe that all schemes fulfill the above-identified most important requirements of fidelity, effectiveness and robustness, except for Guan et al. \cite{guan_reversible_2020} who purposefully give up robustness in favor of reversibility. This is inspired by traditional image integrity: the authors point out that the application of their scheme is not IPP, but integrity authentication, and that all existing watermarking methods are irreversible -- once the watermark is embedded, it cannot be removed to restore the original model without degrading the model's performance.
They argue that irreversible watermarking schemes alter the signature of a model, which could have severe consequences, especially in applications for, e.g., the medical or defense domain.
The fidelity requirement does not apply for Zhang et al.'s method \cite{zhang_model_2020}, since fidelity is not well-defined for generative models. As these output an image (or other complex data), whether a watermarked version of such a model is comparable to the original one requires defining an appropriate similarity measure to determine if two outputs are equivalent.

\subsection{White-box Watermarking}
\label{sec:whitebox}
White-box watermarking requires full access to the model during watermark extraction and verification.

% --- regularizer based - into weights --- fingerprinting potential
The first framework for embedding a watermark into a DNN was proposed by Uchida et al. \cite{uchida_embedding_2017}\footnote{A slightly extended version can be found in \cite{nagai_digital_2018}} in 2017. %TODO: maybe remove footnote and other reference?
They follow the idea of embedding a signature into the model, particularly in the DNN's weights. 
Although it would be possible to directly alter the model's parameters (as in the case of watermarking relational data), this would degrade the model's performance.
The model is trained with a regularizer term, given the signature $\mathbf{b} \in \mathbb{R}^T$, the averaged weights vector $\mathbf{w} \in \mathbb{R}^M$ and a specially crafted \textit{embedding matrix} $\mathbf{M} \in \mathbb{R}^{T \times M}$. The embedding matrix $\mathbf{M}$ can be considered a secret key for the embedding- and extracting processes. The watermark is extracted by applying $\mathbf{M} \in \mathbb{R}^{T \times M}$ to the weights vector $\mathbf{w} \in \mathbb{R}^M$ and then applying a step function. The resulting vector $\mathbf{\tilde b}$ is compared with the signature $\mathbf{b}$, and the bit error rate (BER) is computed. Ownership is proven by thresholding the BER. 

% ---- regularizer based - into pdf ----
Rouhani et al. \cite{rouhani_deepsigns_2019} propose a watermarking framework which proves to be more robust against watermark removal, model modifications, and watermark overwriting than \cite{uchida_embedding_2017}.
This method is regularizer-based and encodes the signature in the PDF of activation maps obtained at different DNN layers, through
an additional regularization term that ensures that selected activations are isolated from others, in order to avoid creating a detectable pattern of alterations.
During the verification process, previously generated trigger images are used as input for the model to then analyze the activations. The scheme can be employed in a white-box- or black-box setting, depending on whether just the output-layer-, or also hidden-layer activations are assumed to be available for watermark verification. 
Note that access to the output activations is not guaranteed in a black-box setting.

% ---- GAN-like network but also regularizer based ---
Wang et al. \cite{wang_riga_2021} show that both previous schemes are vulnerable to watermark detection (cf. \cref{sec:attacks}), as the weight distribution deviated from those of non-watermarked models. The authors claim that this arises from the additive regularization loss function(s). Consequently, they propose a scheme that is particularly robust against detection attacks.
Inspired by the training of GANs, they train a watermarked \textit{target} DNN which is competing against a \textit{detector} DNN that aims to discover if a watermark is embedded.

% --- also regularizer based
Wang et al. \cite{wang_watermarking_2020} follow a similar approach and propose a white-box scheme that makes use of an additional DNN for the watermark embedding process. The target model is trained in parallel with an embedding model, which is kept a secret after the embedding. 
The scheme is regularizer-based, and the watermark is verified by feeding the selected weights into the embedding model and thresholding the output vector. 
They empirically show that their scheme achieves better fidelity, robustness and capacity compared to \cite{uchida_embedding_2017}.

% ---- binarization and compensation mechanism
Feng et al. \cite{feng_watermarking_2020} combine a binarization scheme and an accuracy compensation mechanism to reduce the model's accuracy degradation, which is a result from fine-tuning. They use spread-spectrum modulation on the signature $\mathbf{b}$ and embed it in different layers to reduce the risk of the watermarked weights being set to zero during a pruning attack.
The binarization scheme then transforms the selected weights per layer so that the second norm of the selected weights in one layer remains unchanged, making it harder to discover the embedding position of the watermark.
Finally, they use a regularizer mechanism in the fine-tuning to reduce the impact of watermark embedding on the model's performance.

% ----- Chen: SpecMark:A Spectral Watermarking Framework for IP Protection of Speech Recognition Systems
The first (and so far only) white-box framework for Automatic Speech Recognition (ASR), \textit{SpecMark}, was introduced by Chen et al. \cite{chen_specmark_2020}. They embedded the watermark in the spread spectrum of the ASR model without re-training it. The authors evaluated \textit{SpecMark} on the DeepSpeech model and concluded that it does not have any impact on fidelity.
% -> TODO: hat (noch?) keine Einteilung in IPP_overview

\subsection{Black-box Watermarking}
\label{sec:blackbox}
Black-box watermarking methods need only querying access to the model during watermark extraction and -verification.
Only two of the existing black-box watermarking frameworks \cite{jia_entangled_2021, szyller_dawn_2021} address the second threat model scenario, i.e. the illegal copy (cf. \cref{sec:threatmodel}). All the other methods are not reliably robust against Model Stealing attacks \cite{oliynyk_i_2023}, and therefore primarily address the first case (legal copy).

All frameworks that are defending against the legal copy case utilize \textit{backdooring} via data poisoning (cf., e.g., \cite{gu_badnets_2019}). 
A \textit{backdoor} consists of a so-called \textit{trigger set} of input-output pairs -- which are only known to the backdoor creator (in most cases, the model owner) -- and triggers a behavior that is not predictable by others. We call the input images of the trigger set \textit{trigger images} (sometimes also referred to as \textit{watermarks}).

% --- trigger images figure ----
\begin{figure*}[t]
        \centering
  \subfloat[Out-of-distribution \cite{adi_turning_2018}\label{fig:trigger-a}]{%
       \includegraphics[height = 2.3cm]{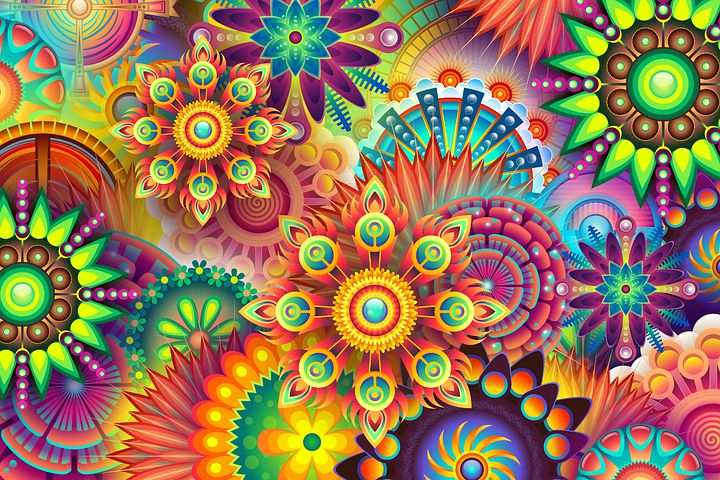}}
        \hfill
    \subfloat[In-distribution \cite{namba_robust_2019}\label{fig:trigger-e}]{%
        \includegraphics[height = 2.3cm]{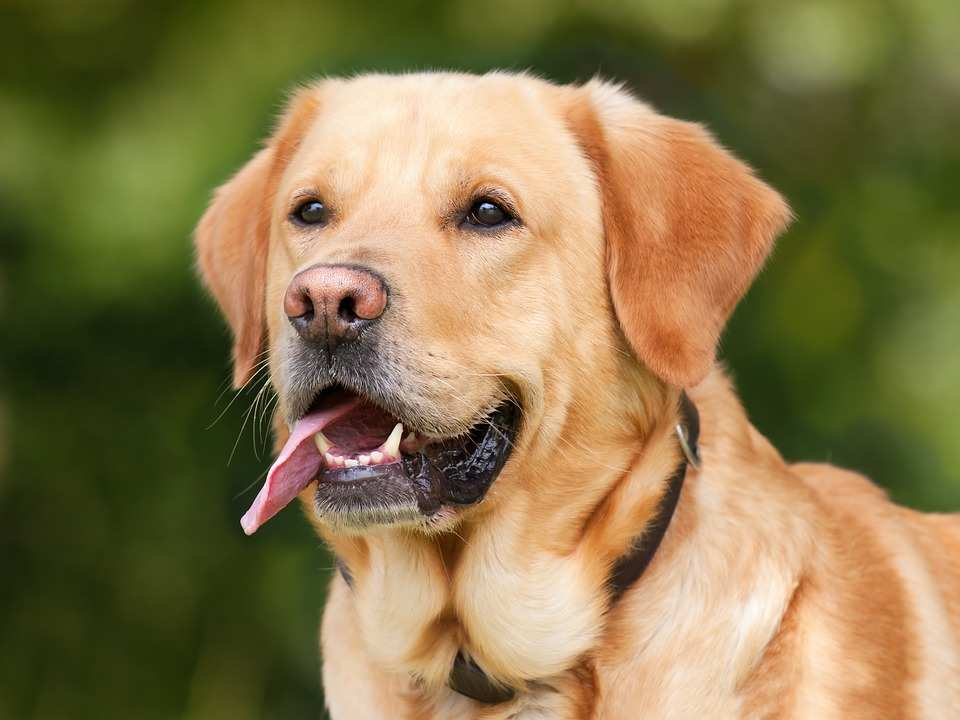}}
    \hfill
  \subfloat[Pattern \cite{zhang_protecting_2018}\label{fig:trigger-b}]{%
       \includegraphics[height = 2.3cm]{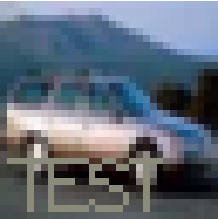}}
    \hfill
  \subfloat[Noise \cite{zhang_protecting_2018}\label{fig:trigger-c}]{%
        \includegraphics[height = 2.3cm]{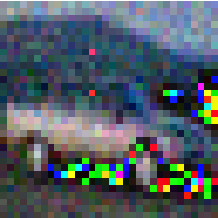}}
            \hfill
    \subfloat[Perturbation \cite{merrer_adversarial_2019}\label{fig:trigger-d}]{%
        \includegraphics[height = 2.3cm]{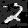}}
        
    \caption{Examples for the various types of trigger images, intentionally labeled as a different class ((a), (b) as "cat", (c), (d) as "airplane", (e) as "9")}
    \label{fig:blackbox}
\end{figure*}
% -------------------------- %

Black-box watermarking methods focus on either creating suitable trigger images (inputs) or the output for the trigger image. Depending on the scheme, different trigger images are used for watermarking: \textit{out-of-distribution} (OOD), \textit{pattern-based}, \textit{noise-based}, \textit{perturbation-based}, and \textit{in-distribution}.
OOD images are completely unrelated to the dataset, for example abstract images in a dataset of handwritten digits.
In-distribution trigger images are taken from the original training dataset and deliberately re-labeled wrongly.
Pattern-based images are derived from the training dataset, e.g. by marking with a pattern such as a logo, text, or other designed patterns. This is comparable to patterns embedded in images for "conventional" data poisoning attacks (\cite{gu_badnets_2019}).
Noise-based images are derived from the training dataset by adding noise (i.e., no systematic pattern), either visible or invisible to the human eye.
Perturbation-based images are slightly perturbed images and lie near the classification boundary, thus, when re-labeled, they force the model to slightly shift its classification boundary, and are inspired by adversarial examples \cite{szegedy_intriguing_2014}.
\cref{fig:blackbox} shows examples for all five types of trigger images. 
Similar to embedding backdoors -- as an attack to reduce the availability or integrity of a model --, the overall objective is that the model will accurately behave on the main classification task, while classifying the trigger images as designated by the owner.

Zhang et al. \cite{zhang_protecting_2018} proposed the first black-box watermarking scheme in 2018 and introduced three types of trigger images: \textit{unrelated} (OOD), \textit{content} (pattern), and \textit{noise}.
Their work was the basis for many subsequent papers.

\subsubsection{Out-of-distribution}
Similar to and shortly after Zhang et al. \cite{zhang_protecting_2018}, Adi et al. \cite{adi_turning_2018} proposed to include abstract images as triggers in the training dataset. Those abstract images are completely unrelated to the main classification task, thus it is highly unlikely that a model that has not seen this data point (i.e., one not watermarked) will label it as the designated class. 

% ---- Quan: Watermarking Deep Neural Networks in Image Processing
One of the first watermarking schemes for image processing models was proposed by Quan et al. \cite{quan_watermarking_2020}. The main difference to classification is that the output is, like the input, an image and not a label -- thus they generating input-output pairs that consist of trigger images and \textit{verification images}. 
They use OOD images (or random noise) as trigger images and create the verification images by applying a simple image processing method to the trigger images (ideally not the one on which the model is being trained). 
The model is then fine-tuned on the union of the original dataset and the trigger set.

% ----- Yang: Effectiveness of Distillation Attack and Countermeasure on Neural Network Watermarking 
Yang et al. \cite{yang_effectiveness_2019} empirically showed that distillation is an effective watermark removal attack.
Therefore, they propose a scheme that they claim to be especially robust against distillation. The main idea is that the watermark information is carried by the predictions of the original training data, whereas the watermark extraction is done by querying an OOD trigger. 
In contrast to \cite{adi_turning_2018} and \cite{zhang_protecting_2018}, the target model is not trained on the union of the original dataset and the trigger set, but only on the original dataset while making use of another model, the \textit{ingrainer model}; this influences the target model by a regularizer term in the loss function. The ingrainer model has the same architecture as the target model and is only trained on the trigger set, with the purpose to overfit the trigger set.

\subsubsection{Pattern}
% ------ Li: Piracy Resistant Watermarks for DNNs
An improved pattern-based technique was proposed by Li et al. \cite{li_piracy_2020}. They show that previous schemes \cite{adi_turning_2018, zhang_protecting_2018} are vulnerable to \textit{ownership piracy} attacks, during which an attacker aims to embed their own watermark into an already watermarked model. The authors propose a scheme that is especially robust against such attacks using so-called \textit{dual embedding}:
the model is trained to classify (i) data with a pre-defined binary pattern correctly, i.e., \textit{null embedding}, and (ii) data with an inverted pattern (binary bits are switched) incorrectly, i.e., \textit{true embedding}.
They observe that null embedding does not degrade the model's accuracy if the number of pixels in the pattern is sufficiently small.
Furthermore, they evaluated the robustness against Model Stealing attacks, and concluded that with out-of-distribution data, the attacker would need significantly more input data to reach similar accuracy.

% ---- Guo: Watermarking Deep Neural Networks for Embedded Systems
Guo et al. \cite{guo_watermarking_2018} proposed to embed a pattern into the trigger images that can be clearly associated with the model owner's signature, e.g., a logo. The pattern should be embedded with little visibility so that an unmarked model would still classify the trigger images according to its original labels. 

% ---- Evolutionary trigger set generation --- pattern based
As an improvement to \cite{guo_watermarking_2018}, Guo et al. \cite{guo_evolutionary_2019} proposed an evolutionary algorithm-based method to generate and position trigger patterns. 
Their algorithm is based on Differential Evolution \cite{storn_differential_1997}, %TODO: remove reference if we need reference space
an evolutionary algorithm and metaheuristic that searches for solutions to an optimization problem.
Using this trigger pattern generation, they demonstrated an improvement in integrity and robustness.

\subsubsection{Noise}
\begin{figure}[t]
    \centering
    \includegraphics[width = 0.45 \textwidth]{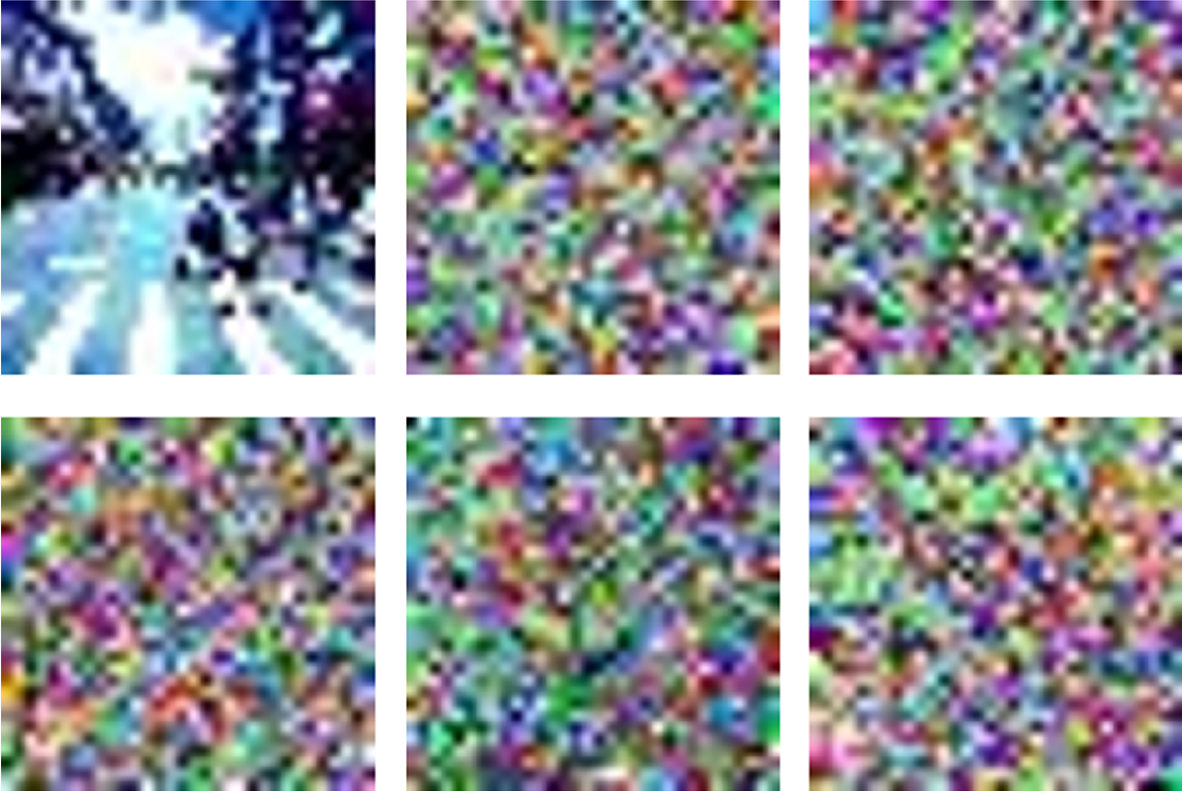}
    \caption{The upper left image is the initial image, and the following five are trigger images resulting from a hash chain \cite{zhu_secure_2020}.}
    \label{fig:zhu}
\end{figure}

% ---- Secure NN WM protocol --- noise
Zhu et al. \cite{zhu_secure_2020} proposed a watermarking scheme to defend especially against overwriting. They used one-way hash functions to generate both the trigger image and -label.
The framework takes an initial image and creates a hash chain of trigger images, as shown in \cref{fig:zhu}. They showed experimentally that their proposed scheme is robust against overwriting even if the attacker knows the trigger set generation algorithm.

\subsubsection{Perturbation}
\begin{figure}[t]
    \centering
    \hspace*{\fill}
  \subfloat[\label{fig:merrer-a}]{\includegraphics[width = 0.215 \textwidth]{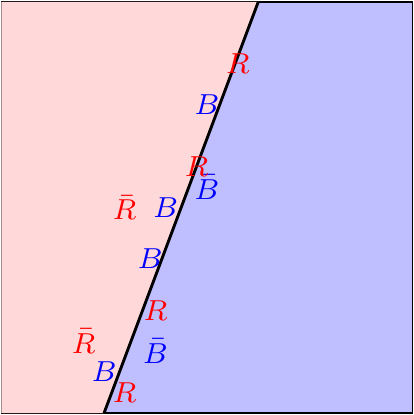}}
    \quad
  \subfloat[\label{fig:merrer-b}]{\includegraphics[width = 0.215 \textwidth]{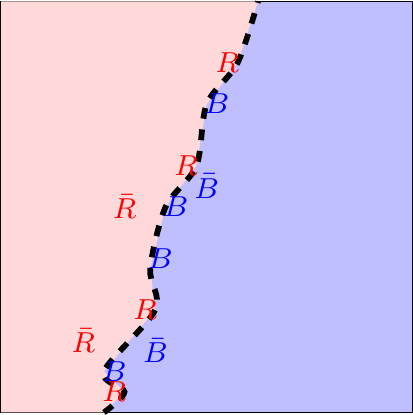}}
    \hspace*{\fill}
    \caption{(a) The data points will be divided into "true adversaries" ($R$ and $B$) and "false adversaries" ($\bar{R}$ and $\bar{B}$). The label for the true adversaries is changed, the label for the false adversaries stays unchanged. (b) After fine-tuning the decision boundary has changed. \cite{merrer_adversarial_2019}}
    \label{fig:merrer}
\end{figure}

% ---- adversarial frontier stitching
The goal of Merrer et al. \cite{merrer_adversarial_2019} is to slightly shift the decision boundary of the model. This is achieved by generating adversarial examples \cite{szegedy_intriguing_2014} for images close to the boundary, and changing the assigned class label.
After fine-tuning the model, the decision boundary is adapted. An illustration of this decision boundary shifting is given in \cref{fig:merrer}.

% ----- GAN-based technique
Li et al. \cite{li_how_2019} especially address evasion attacks. They proposed a framework closely related to the idea of GANs, and use three DNNs: encoder, discriminator, and target model. The encoder takes the original image and aims to embed a logo into the image in a way that the difference is imperceptible. The resulting trigger images are fed into the discriminator -- together with the original image -- to evaluate the encoder's success.
A difference in the original- and trigger images is essential for the effectiveness of the embedded watermark -- the larger the difference, the better. However, the smaller the difference, the better the protection against evasion attacks -- the authors thus specifically aim to address this trade-off.

\subsubsection{In-distribution}
Namba et al. \cite{namba_robust_2019} proposed an attack called \textit{query modification} to invalidate the watermark, which exploits the fact that trigger images differ from original training images (cf. \cref{sec:attacks}).
They then developed a scheme that is more robust especially against query modification, but also model modifications like fine-tuning and model compression (e.g., pruning). 
They suggest to use trigger images that are selected from the training sample distribution.
Although the trigger images are undetectable, the model is more likely to overfit the (on purpose) wrongly labeled triggers and, thus, more susceptible to removal attacks via, e.g., pruning. 
They want to counter this pruning by ensuring that the predictions do not depend on a large number of small model parameters which would likely be pruned. Therefore, the model is first trained as usual with the original training set. Then the watermark is embedded through exponentially weighting the parameters and training the model on a combination of the original dataset with the trigger set, which enforces the predictions to depend on a small number of large parameters.

\subsubsection{Label}
The papers in this category focus on the label for the triggers. Hence, the choice of trigger images is secondary.

% ----- Zhong: Protecting IP of Deep Neural Networks with Watermarking: A New Label Helps --- does not rely on specific choice of pattern
Zhong et al. \cite{zhong_protecting_2020} propose to label the trigger images with a completely new label rather than assigning an existing one, so that the watermark embedding has only little impact on the original classification boundaries.
Any pattern-based trigger image can be used in this context. They empirically compare their work to \cite{zhang_protecting_2018} and show that the proposed scheme achieves a zero false-positive rate, i.e., excellent integrity, and is more robust against fine-tuning and model compression.

Zhang et al. \cite{zhang_deeptrigger_2020} observe that trigger images are frequently created in a systematic way, which makes it is easier for an attacker to re-create them.
Therefore, they propose to include unpredictability in the labels for the triggers. They use a chaos-based labeling scheme that ensures that an attacker cannot produce a valid trigger set, even if they know the pattern.

\subsubsection{Output vector}
% ---- Chen: BlackMarks 
% category: output activations
The approaches in this category focus on embedding information in the output vector.
Chen et al. \cite{chen_blackmarks_2019} propose the watermarking framework \textit{BlackMarks} which 
encodes the signature within the distribution of the output activations. 
To encode the class predictions, the authors design a scheme that maps the class predictions to bits, i.e., by clustering the original classes into two categories represented by bit 0 and bit 1. The trigger images are created as follows: take an image from cluster '0', create an adversarial example so that it would be predicted with a class belonging to cluster '1' and finally label it with a uniformly randomly chosen class from cluster '0'.
Trigger images for bit 1 are created vice-versa. The watermark is extracted by querying the trigger images and encoding each class to a binary value, which should result in the owner's binary signature.

% ----- Xu: Identity Bracelets --- does not rely on specific choice of trigger images
Similarily, Xu et al. \cite{xu_identity_2020} proposed, a watermarking scheme that carries the watermark information within the output activations. The trigger pairs consist of a trigger image and a serial number (SN), which is encoded in the model's probabilities.

\subsection{Countering Model Stealing} \label{sec:wm:countering_extraction}
% --- Jia: EWE
Only a few schemes address the second threat scenario in \cref{sec:threatmodel}, i.e., robustness against Model Stealing attacks \cite{oliynyk_i_2023}. 
First, Jia et al. \cite{jia_entangled_2021} proposed a scheme called \textit{Entangled Watermark Embedding} (EWE).
The main idea is to create a watermarked model that is not specialized into "sub-models", where one part of the model has learned the main classification task and the other the watermark detection (which is normally lost during a Model Stealing attack). This is achieved through a regularizer ensuring that the trigger images lead to similar activation patterns as the original images. Thus, both trigger- and original images cause a similar behavior of the model, thereby increasing the robustness against model stealing.  

% ---- Szyller: DAWN
Szyller et al. \cite{szyller_dawn_2021} proposed the framework \textit{Dynamic Adversarial Watermarking of Neural Networks} (DAWN) which does not embed a signature into the target model itself, but dynamically returns wrong classes from the API service for a fraction of queries to mark an adversary model created via a Model Stealing attack.
It is worth noting that the scheme is not able to differentiate between an attacker and a benign client -- all clients obtain a fraction of wrong predictions, and it is ensured that the same query always returns the same output (correct or modified) to avoid simple collusion attacks.
This approach thus realizes \circled{4} in \cref{fig:watermarking-ml-process}.

% ---- Zhang: Model Watermarking for Image Processing Networks
Zhang et al. \cite{zhang_model_2020} and Wu et al. \cite{wu_watermarking_2021} proposed, independently of each other, an approach similar to \cite{szyller_dawn_2021}, as they are hiding an invisible watermark in the outputs of the image processing model, but for \textit{all} outputs. When an attacker trains a new (surrogate) model on the input-output pairs of the original model, the watermark will be learned as well and can be verified via black-box access (cf. \circled{4} in \cref{fig:watermarking-ml-process}). One difference to \cite{szyller_dawn_2021} is that in the case of an image-processing model, the output is another image, and thus there is more space to embed the watermark in. Neither papers explicitly address Model Stealing attacks, but we believe they are a suitable defense.

\subsection{Watermarking for specific ML settings}
% ------ Tekgul (was: Atli): WAFFLE: Watermarking in Federated Learning
Existing watermarking schemes are not suitable for Federated Learning (FL), as pointed out by Tekgul et al. \cite{tekgul_waffle_2021}. 
Embedding a watermark in such a setting is different, because the model owner has no access to training data, and the training is performed in parallel by several clients.
Tekgul et al. propose to include an independent and trusted third party between the model owner and the clients, which will embed a backdoor-based and black-box watermark into the model at every aggregation step.
Furthermore, they propose a specific noise pattern for the triggers.

% ------ Skripniuk: Black-Box Watermarking for Generative Adversarial Networks
Yu et al. \cite{yu_artificial_2021} proposed the first watermarking scheme specially crafted for GANs. Previously existing watermarking schemes were limited to DNNs that map from images to classes, and thus could not be transferred to GANs. Yu et al. watermarked the input images and then transferred these images to the GAN model. Thereby, the image steganography system, which consists of an encoder and decoder, has first to be trained and, subsequently, \textit{all} the training data -- together with a secret watermark -- is fed to the encoder, resulting in watermarked data. The watermarked data is then used to train the GAN model. For verification, the model owner only needs an output image of the GAN, and applies the decoder on it to compare the result with the secret watermark. Thus, the proposed scheme needs only black-box access for verification.

\subsection{Watermarking as part of an IPP Workflow}\label{sec:watermarking-workflow}
Besides the above-mentioned watermarking techniques as reactive methods, further approaches have been proposed to prove ownership of IP in other domains, e.g., using ledgers such as blockchains to deposit the digital object (or a signature thereof) together with the owners identity~\cite{wust_you_2018,savelyev_copyright_2018}. 
We note that such mechanisms could be one option to prove ownership of an ML model, if white-box access to the model exists and the model itself or its signature can thus be compared to the ledger's entries. 
To be practical, such schemes would need to be robust to small changes in the model parameters -- an attacker could perform those changes at little extra loss due to general model robustness, while the changes would invalidate ownership claims, as models and signatures would not match anymore; hence, approaches such as fuzzy hashing might be a suitable solution.
Additionally, as discussed in \Cref{sec:watermarking}, white-box access to a suspected pirated model is unrealistic, as ML models can be exploited and monetized by clients without the need for this type of access.

For black-box methods, which have superseded white-box approaches for reasons outlined in \Cref{sec:watermarking}, access to the model itself is not available, and, therefore, no signature can be computed for comparison with a deposited model.
To support ownership verification, the aim of the legitimate model owner is to prove that they have knowledge of the trigger set pairs (inputs and expected outputs), which can be seen as a form of challenge-response. 
As other users of the model could also search for inputs with unexpected outputs (also ex-post with black-box access) and claim these to be valid evidence proving ownership, it is of interest to owners to prove their knowledge at a specific (and earlier) \textit{point of time}. 
Depositing signatures of these trigger pairs with a trusted authority (such as a notary) or a distributed ledger (such as blockchain) can provide a trusted time stamp that can be used in the verification process.

\subsection{Length of Watermarks and Complexity}
In most cases, the length of the watermark is either determined by the number of parameters changed in white-box watermarking, or the number of trigger images used in black-box approaches.
Regarding complexity, the embedding time heavily depends on the choice of either training the model from scratch or fine-tuning it, as fine-tuning creates additional overhead. 
Regarding extraction time, there are no major differences between the methods, as all of them either score trigger images or transform the model’s parameters in order to extract the watermark -- they are thus all dependent on the (i) watermark’s length as well as (ii) the prediction time in case of black-box watermarking, or the number of neurons on the chosen layer in white-box watermarking.

\section{Fingerprinting of ML Models} \label{sec:fingerprinting}

A model owner might be selling their ML model to different customers, but the model gets illegally re-distributed by one of them. The owner would then like to gather evidence on the leak; therefore, they could embed fingerprints in the ML model before selling the product in order to trace back a malicious user if needed. We can think of fingerprinting as a user-level extension of watermarking.
At the time of performing this systematization, fingerprinting for ML models was not extensively discussed, with only three papers published.

Note that there is another definition of fingerprinting: a (unique) identifier for an object (either hardware, software, or a combination thereof) is generally referred to as a "fingerprint", e.g., such as in browser fingerprinting~\cite{eckersley_how_2010} or device fingerprinting~\cite{kohno_remote_2005}. 
The application scenario for employing these techniques is often to track devices (resp. their users). 
Also, this use of fingerprinting is an inherent property of the object, and not the result of an active embedding process.
Given that this context differs from what we considered so far, we call this form \textit{fingerprinting as unique identification}.

\subsection{Fingerprinting as User-specific Watermark}
\label{sec:fingerprinting:user-specific-wm}

% regularizer based
Chen et al. \cite{chen_deepmarks_2019} propose \textit{DeepMarks}, a white-box fingerprinting framework that is able to embed unique fingerprints. The verification process not only detects the malicious user, but also if multiple -- and if so which -- users collaborated in order to remove the watermark. The embedding process works similar to DeepSigns \cite{rouhani_deepsigns_2019}. The authors propose to assign a unique binary vector (fingerprint) to each user and embed the fingerprint information in the PDF of the weights before distributing the models to the users.

Although \textit{DeepMarks} is the only paper especially considering fingerprinting, we believe that a couple of the watermarking schemes introduced above can be extended to fingerprinting. To name a few, \cite{uchida_embedding_2017} embeds a unique signature into the weights of the DNN, \cite{li_how_2019} embeds a unique logo into the trigger images and \cite{guo_watermarking_2018} generates unique trigger images based on a signature. 
All of them could embed user-specific watermarks. Moreover, \cite{xu_identity_2020} relies on serial numbers that can be created in indefinitely many ways, assigning each to a user.

\subsection{Fingerprinting as Unique Model Identifier}
Cao et al. \cite{cao_ipguard_2021} proposed a framework to obtain a unique identifier of DNNs.
As two different models likely have different classification boundaries, they suggest to "fingerprint" this boundary. The authors identify so-called "fingerprinting data points" that lie near the model's classification boundary. Since the points lie \textit{near} the classification boundary rather than \textit{on} it, the authors claim robustness against model modifications and uniqueness of the fingerprint.

% perturbation
Zhao et al. \cite{zhao_afa_2019} and Lukas et al. \cite{lukas_deep_2021} modified this idea of fingerprinting as unique identification. Both propose a scheme in which the adversary model -- created through applying modifications to the target model -- has the same fingerprint as the target model. Both introduced a novel algorithm for creating transferable adversarial examples (see, e.g., \cite{liu_delving_2017}). 
In \cref{sec:blackbox}, we described how black-box watermarking methods use perturbation-based trigger images (i.e., adversarial examples) which are used during training so that the models learn how to (purposefully) misclassify them.
In the context of fingerprinting as unique model identifier, the authors want to instead create an adversarial example from an \textit{already trained} ML model. 
The key aspect is that these generated images are not only adversarial examples for the target model, but also for the adversary model, i.e., they are transferable. This fits our first threat model in \cref{sec:threatmodel}.

\section{Access Control and Other Proactive IPP} \label{sec:otherIPP}
In this section, we analyze proactive IPP methods. These are orthogonal to and go further than ownership verification.

The most prominent type of methods tries to \textit{prevent unauthorized access} to a trained neural network. This is achieved by rendering the model useless to an unauthorized user, even if this user manages to obtain a full and exact copy of the model. 
Most methods employ obfuscation and/or encryption, which can only be overcome with a matching \textit{secret}.

There are diverging viewpoints on which assets of an ML model are most important to protect. The majority of literature argues that it's the \textit{learned model parameters}, as (i) learning requires large amounts of (expensive) training data, expertise with training the model, and computing resources, and (ii) in many cases, standard, well-known architectures (such as GoogLeNet/Inception \cite{szegedy_going_2015}, ResNet, etc.) are employed -- the architectures are not secret, but the models need to be (re-)trained to fit the domain.
However, other works (e.g., \cite{xu_deepobfuscation_2018}) highlight the fact that if a custom architecture is developed, then the resulting structure is actually the asset to protect.

Analogous to ownership verification (cf. \cref{tab:requirement}), access control mechanisms should primarily fulfill the following requirements.
%\begin{itemize}
    %\item 
    \textbf{Fidelity}: the model should maintain accuracy after applying the pro-active defense.\\ 
    %\item 
    \textbf{Robustness}: the access control should resist a designated class of transformations, including malicious model modifications.\\
    %\item 
    \textbf{Efficiency}: the impact of the access control mechanism on the time for prediction (and to some extent also for training); this is more relevant than its counterpart for ownership verification (watermarking), as it might affect normal operation efficiency.\\
    %\item 
    \textbf{Protection effectiveness}: a model that is used without proper authorization should incur a significant \textbf{degradation} in prediction correctness, so that the value for the attacker is diminishing or even non-existent. For example, the authors of \cite{lin_chaotic_2020} set a loss of 20 percent points in effectiveness as goal to render the model useless.
%\end{itemize}

The choice of protection scheme also depends on the type of asset to be protected.
We can, in general, distinguish the following approaches (cf. \cref{fig:overview}):
\begin{enumerate}
    \item \textbf{Obfuscating} the model structure
    \item \textbf{Modifying the input}, e.g., by encryption or permutation
    \item \textbf{Encrypting} (parts of) the model, i.e., the weights
    \item \textbf{Modifying} the model \textbf{structure}, e.g., by adding layers
\end{enumerate}

\input{tab-summary-req-accessControl}
We provide an overview of the pro-active IPP schemes considered in this paper in \cref{tab:summary-req-accessControl}, where we indicate whether they are meeting the above-mentioned requirements.
While fidelity and protection effectiveness are discussed or demonstrated by almost all works, we can observe that only a selected number of papers is doing this for robustness against attacks as well as for the efficiency of their scheme.

One aspect common to most access control schemes is that they require the authorized user to posses some form of \textit{secret}, e.g., a key or a token.
Most works, however, do not discuss aspects of management and revocation of these secrets.
While those aspects are somewhat orthogonal to the access control mechanism itself, they are of significant importance, as the most commonly discussed scenario is that the models are deployed in the customer's infrastructure, or, e.g., in an embedded device. 
To some extent, this lack of holistically considering pro-active IP protection mechanisms is comparable to using watermarking without considering a proof of existence of the trigger sets, (as discussed in \Cref{sec:watermarking-workflow}).

\subsection{Model Architecture Protection}
% DeepObfuscation
Xu et al. \cite{xu_deepobfuscation_2018} propose a scheme to protect the structure of a CNN. They argue that (i) being able to pirate an architecture, even without trained weights, is a major incentive for adversaries, and (ii) the most important part of the architecture is the feature extraction in the early layers -- and not the fully connected layer(s) at the end.
They thus propose to \textit{obfuscate} these layers of the architecture through iteratively replacing complex processing \textit{blocks} of a trained CNN, such as an inception module \cite{szegedy_going_2015}, by a small number of sequentially aligned convolutions (the \textit{simulation network}).
For the simpler, shallow structure to learn these modules effectively, they are using teacher-student network approaches, receiving as ground truth the class label \textbf{and} the output of the feature extraction block.
The resulting network does not suffer a noticeable effectiveness loss and is in most cases more efficient. The shallow structure is, however, not capable to learn a new, similarly complex task from scratch and, thus, of only limited utility to an attacker.

\subsection{Model Parameter Protection}

% Weights encryption
\subsubsection{Parameter Encryption and Obfuscation}
Gomez et al. \cite{gomez_security_2019} proposed a scheme utilizing homomorphic encryption (HE), which allows to compute certain operations directly on encrypted data. 
Prediction is performed by encrypting the input to the homomorphically encrypted layers, and decrypting the resulting output, using asymmetric key pairs. Encryption is \textit{limited to parts} of the model due to the huge runtime overhead incurred by HE.
While the authors argue that the last layers they train are the most valuable -- and, thus, get encrypted -- others (e.g., Fan et al. \cite{fan_rethinking_2019}) argue that the first layers, which are responsible for feature extraction, are the important asset -- especially if the training dataset is not public and the feature extractors differ from those of benchmark datasets.

Chakraborty et al. \cite{chakraborty_hardware-assisted_2020} introduce a \textit{hardware-protected} NN (HPNN). 
In their threat model, they assume that an attacker obtains the white-box model, and to host their own (public) service or to use it in a private environment. 
They argue that encrypting thec whole NN will lead to prohibitive runtime overhead for predictions, which are, however, often required in (near) real-time. 
They obfuscate the learned parameters through a technique they call \textit{locking} -- 
i.e., making some neurons in the network dependent on a secret key that determines the sign of the value of the linear function in that neuron.
The model can thus be openly distributed, as to correctly use it, a secret key needs to be available in a trusted hardware (root of trust, such as a Trusted Platform Module (TPM)). 
The model needs to be trained with a modification of the backpropagation algorithm to be key-dependent, but this does neither affect fidelity nor the model's ability to learn the relationship between inputs and outputs.
The authors argue that a hardware solution entails stronger security guarantees and less performance overhead.

Alam et al. \cite{alam_nn-lock_2022} propose using a key-scheduling algorithm to create a series of keys, one for each model parameter of a DNN. After a standard training algorithm, the owner encrypts the parameters using a substitution-box (S-box).
At prediction time, a legitimate user owning a key uses the same key-scheduling algorithm to create decryption keys for each query.
%They demonstrate fidelity and protection effectiveness.

Lin et al. \cite{lin_chaotic_2020} use chaotic encryption to obfuscate the model parameter \textit{positions} without changing the weights' distribution, thus making detection of this scheme more difficult.
Legitimate users need a key to determine the positions of the output cells of kernel (matrix) operations, otherwise these outputs will be in a wrong sequence and the prediction correctness will deteriorate.
The authors argue that the decryption is (i) fast enough, as only a few layers need to be encrypted, and (ii) secure enough, as the decryption can be performed independently for each layer and on the chip, rendering memory attacks impossible.
One challenge is selecting the most effective combination of layers to encrypt, as, depending on the architecture, not all layers provide the same protection.

Motivated by serial number (SN) verification in software products, Tang et al. \cite{tang_deep_2020} proposed a scheme for DNNs in which the user has to posses a valid serial number. The SN could be a secret combination with the model owner's identity and can therefore be also used for ownership verification. The embedding is done by a teacher-student framework where the teacher network learns the classification task and the student network is distilled from the teacher network, with an additional loss that ensures that the SN is embedded.

\subsubsection{Input Obfuscation}
% Input based

Chen et al. \cite{chen_protect_2018} propose a scheme using a \textit{transformation module} that pre-processes inputs in a secret way before passing them to the prediction module, which needs no further protection -- inputs that are not pre-processed correctly will deteriorate accuracy.
For the transformation module, the authors invert the idea of adversarial examples: adding an adversarial perturbation specific for each input, so that the model to \textit{correctly} classifies them. The module thus acts as a kind-of decryption module and is intended to run in a TPM.
In their most successful approach, the transformation module is implemented as a CNN and trained together with the prediction module using specific regularizers. 

AprilPyone et al. \cite{aprilpyone_training_2020} propose a scheme in which \textit{inputs} to the model are perturbed in a specific (deterministic) manner through block-wise pixel shuffling before they are fed to the model training or prediction phase.
The perturbation is based on a secret that the rightful user possesses -- not knowing this key will results in distortion in the spatial arrangement that will render the model ineffective.

\subsubsection{Structure Modification}
% Modifying network structure
Fan et al. \cite{fan_rethinking_2019} proposed \textit{passport layers}, a scheme of inserting additional layers into the network. These layers are added after the convolutional layers and perform a scaling operation, the parameters of which are derived from the secret, called \textit{passport}. These are generated, e.g., based on a given set of input images and the values in the feature maps which result when passing them through a trained model of the same architecture.
% TODO: maybe not here? maybe in attacks?
The author's motivation is not primarily access control, but adding a kind of ``second factor'' to ownership verification. They argue that it is easy to forge watermarks for a given model and thus have a false ownership claim, e.g. against \cite{uchida_embedding_2017} or \cite{adi_turning_2018}, as trigger sets based on adversarial examples do not depend on the input data and thus can be obtained from a trained model alone (see \cref{sec:attacks}).
To additionally demonstrate high fidelity of the model, is, however, only possible when having authorized access.
This additional step for ownership verification could be provided by most of the other access control schemes presented in this section.

Sun et al. \cite{sun_convolutional_2018} show -- using the example of a LeNet-5 CNN -- that adapting the activation function of the convolutional layers to be dependent on a random number only known to the legitimate user can provide effective protection.

% ------ Lim: Protect, Show, Attend and Tell: Empower Image Captioning Model with Ownership Protection
Lim et al. \cite{lim_protect_2022} are the first to propose an access control scheme for a recurrent neural network (RNN). Specifically, they consider an image captioning model producing a text sequence, implemented as a simplified variant of the \textit{Show, Attend and Tell} model \cite{xu_show_2015}. The proposed framework is similar to \cite{fan_rethinking_2019}, while not embedding the verification information (the owner's key) into the model weights, but into the signs of the the hidden states of the RNN.
During model inference time, the key is required as input to the model by an element-wise combination with the input data.

\subsection{Unrobust Models as IPP}
% Mimosanet
Szentannai et al. \cite{szentannai_preventing_2020} observe that published DNNs are useful as they produce robust predictions even with minor perturbations of the parameters. They thus propose a proactive defense mechanism that renders the model sensitive and fragile through applying transformations that add neurons on any hidden layer of the model. 
These neurons decompose previously existing neurons in such a way that the mapping between its preceding and subsequent layer is kept, but weights of existing neurons on the modified layer are divided and, thus, more susceptible to small changes in values.
As a consequence, even minor modifications of the model parameters, caused by, e.g., fine-tuning, will drastically alter the predictions; subsequently, adversaries cannot utilize the model in a transfer learning setting.
In order to make it difficult to spot these additional neurons, so-called "deceptive neurons", which bear no other functionality, are added as decoys.

\section{Attacks on IPP mechanisms}
\label{sec:attacks}

%TODO: make this more general, not only watermarking...
If an attacker knows or suspects that a model is protected, they could try to change the model in order to remove or overwrite the protection. 
Regarding watermarking, most authors claim that their techniques are robust against various model modifications like \textbf{fine-tuning} -- re-training the model with new data -- , and \textbf{model compression} or parameter pruning -- setting small parameter values to zero \cite{han_deep_2016,zhu_prune_2018}. %TODO: remove the second reference?
Still, several attacks that are aiming to remove, overwrite, detect, or invalidate state-of-the-art watermarking schemes have been proposed.
We will analyze those in the following.

\input{tab-attacks-transpose-new}

In \cref{tab:attacks}, we summarize the attacks on watermarking schemes.
Each line corresponds to an attack and each column to a (type of) watermarking scheme. The table shows which attack defeats which kind of watermarking. We list only schemes that were proven to be successfully defeated -- missing schemes in the table do not imply strong robustness. We can see that an attack usually addresses either white-box- or black-box watermarking schemes. The four trigger image types -- OOD-, pattern-, noise- and perturbation-based -- seem to be defeated in a similar way. In-distribution watermarks are more difficult to detect or remove, probably because of the fact that they do not differ from the original training data distribution.
 
\subsection{Watermark Overwriting}
Li et al. \cite{li_piracy_2020} showed that some schemes \cite{adi_turning_2018, zhang_protecting_2018} are vulnerable to watermark overwriting (they call this "ownership piracy").
They applied the schemes to four image classification tasks, and assumed that an attacker would have access to around 10\% of the original training data. They then showed that an attacker could successfully embed its own watermark by fine-tuning the model with data adapted to this watermark.
  
\subsection{Watermark Detection}
Several attacks exploit the fact that a watermarked model actually learns two tasks: the main classification task and the watermark extraction task. 
% ---- Wang: Attacks on Digital Watermarks for Deep Neural Networks
Wang et al. \cite{wang_attacks_2019} revealed vulnerabilities against watermark detection, when they observed that in regularizer-based watermarking methods like \cite{uchida_embedding_2017}, the variance of the distribution of model parameters (they call this \textit{weights variance}) increases during watermark embedding.

Wang et al. \cite{wang_riga_2021} showed that regularizer-based watermarking schemes are vulnerable to watermark detection through the use of a property inference attack \cite{ganju_property_2018}.
Knowing the embedding algorithm, they trained a set of models with similar architecture and similar data (so-called "shadow models"), some of which will be watermarked, others not.
From these models, they extract weights as representative features, and subsequently train a model on these features to distinguish between watermarked and not-watermarked models.
Similarily, Shafieinejad et al. \cite{shafieinejad_robustness_2021} also propose to use property inference for watermark detection.

\subsection{Watermark Removal}
% ---- Wang: Attacks on Digital Watermarks for Deep Neural Networks

Wang et al. \cite{wang_attacks_2019} further removed watermarks by embedding additional watermarks into the model, following the embedding scheme in \cite{uchida_embedding_2017}. Since every additional watermark might increase the weights variance, they propose to lower it by adding an $L_2$ regularizer. Following this procedure, the authors show that the old watermark cannot be extracted, thus the model owner cannot claim ownership. It should be noted that although additional watermarks are embedded into the model, the main objective is to "neutralize" the old watermark rather than to use the new watermarks to claim ownership.

% ----- Shafieinejad: On the Robustness of the Backdoor-based Watermarking in Deep Neural Networks -----
% detection: property inference attack
Shafieinejad et al. \cite{shafieinejad_robustness_2021} analyzed the robustness of backdoor-based watermarking schemes. In particular, they propose a Model Stealing attack that trains a substitute model (\cite{oliynyk_i_2023}).
This is performed by querying the original model with a public dataset from the same domain, and using the resulting label to train their own model. As the public dataset contains none of the trigger images, the watermark is "lost" in the process. 
We want to point out that most of the techniques, as per their design, are not robust against Model Stealing attacks \cite{oliynyk_i_2023}, as pointed out by Mosafi et al. \cite{mosafi_stealing_2019}. 
Exceptions like \textit{EWE} \cite{jia_entangled_2021} and \textit{DAWN} \cite{szyller_dawn_2021} are desribed in \cref{sec:wm:countering_extraction}.

% ----- Liu: Removing Backdoor-based WM in Neural Networks with Limited Data -----
Liu et al. \cite{liu_removing_2021} proposed \textit{WILD}, a framework against backdoor-based watermark techniques embedded via fine-tuning.
They argue that it is hard for attackers to collect the required amount of within-domain, unlabeled data for the attack in \cite{shafieinejad_robustness_2021}, but that using out-of-domain data impacts the effectiveness of the substitute model too much.
Their method requires less data, as they augment it by Random Erasing \cite{zhong_random_2020}, i.e., removing random segments from the input images.
This augmented data alone is, however, not enough to remove a watermark via fine-tuning, due to the high diversity of potential watermarks. 
The authors note that backdoor patterns are mostly learned by the high-level feature spaces produced by the convolutional layers, and not by the fully connected layers.
They thus additionally add a regularizer term that ensures a minimal distance in distribution between the high-level feature space of the augmented- and the clean dataset during fine-tuning, so that a backdoor pattern could not be learned.
The authors reveal that it is more difficult to remove OOD-, compared to pattern- and noise-based watermarks.

% ----- Guo: The Hidden Vulnerability of Watermarking for Deep Neural Networks -----
Guo et al.'s removal attack \cite{guo_fine-tuning_2021} covers two aspects: (i) input data pre-processing consisting of pixel-level alterations such as embedding imperceptible patterns and spatial-level transformation such as affine and elastic transformation, aiming at making the trigger image unrecognizable by the model; and (ii) fine-tuning, with data that can be unlabeled and from a different distribution. The second step aims at restoring the accuracy of the model on normal samples, which might suffer from the input data pre-processing. Using the watermarked model as an oracle to obtain labels, these input samples are then pre-processed in the same manner and used for fine-tuning the model.
The authors empirically show that their watermark removal attack can remove various types of watermarks without knowledge about the watermark embedding or labeled training samples.

% ----- Chen: REFIT: a Unified Watermark Removal Framework for Deep Learning Systems with Limited Data -----
Chen et al. \cite{chen_refit_2021}\footnote{Previous version in \cite{chen_leveraging_2019}} propose \textit{REFIT}, a watermark removal framework based on fine-tuning. 
The basis of their work is the phenomenon of \textit{catastrophic forgetting} \cite{goodfellow_empirical_2015}, which means that models which are trained on a series of tasks can easily forget the previously learned tasks. Their attack model assumes that the attacker has no knowledge on neither the watermark nor the watermarking scheme, and has limited data for fine-tuning.
They first show that in case the training data is known, the watermark can be removed by fine-tuning when choosing the learning rate appropriately. In order to adapt to having only limited data that do not come from the original dataset, the authors include two techniques: (i) elastic weight consolidation (EWC), and (ii) augmentation with unlabeled data (AU). 
EWC slows down the learning of parameters that are important for previously trained tasks, in particular the main classification task, via adding a regularizer term to the loss function. 
AU on the other hand increases the number of in-distribution, labeled fine-tuning data. To this end, unlabeled data is obtained via web scraping and labeled by the pre-trained model. In most cases, the model labels the data according to their true classes, since the model has not seen the data before, and the watermarked model was trained to fulfill the integrity requirement. 
The authors showed that the proposed framework successfully removes the watermark from various state-of-the-art watermarking schemes without degrading the test accuracy.

% ----- Aiken: Neural Network Laundering: Removing Black-Box Backdoor Watermarks from Deep Neural Networks -----
Aiken et al. \cite{aiken_neural_2021} proposed a method for watermark removal based on previous backdoor removal attacks \cite{wang_neural_2019, liu_fine-pruning_2018}, assuming an attacker with a small (less than 1\%) amount of original training data.
Their technique involves three steps: First, they reconstruct the perturbations (backdoor patterns) that are required to flip a sample to the other class, using the method from \cite{wang_neural_2019}. 
Secondly, they superimpose the pattern on their clean training data to identify neurons that are responsible for recognizing the backdoored images, similar to \cite{liu_fine-pruning_2018}. The weights incoming to these neurons are then set so that they produce zero activation. Finally, the model is fine-tuned on the clean and backdoored training data, while labeling the backdoored training data to the class that is least likely to be watermarked, which prevents re-appearance of the neurons that were reset in the previous step.
The authors showed that their technique defeats the watermarking schemes \cite{zhang_protecting_2018} and \cite{adi_turning_2018} by effectively removing neurons or channels in the DNN's layers which contribute to the classification of trigger images.

% ---- Yang: Effectiveness of Distillation Attack and Countermeasure on Neural Network Watermarking -- attack %yang_effectiveness_2019

\subsection{Watermark Invalidation}
Watermark invalidation does not aim to remove the watermark, but finds a way to render it useless.

% ----- Hitaj: Evasion attacks against watermarking techniques found in MLaaS systems
Hitaj et al. \cite{hitaj_evasion_2019} proposed two such attacks: an \textit{ensemble attack} and a \textit{detector attack}. 
The ensemble attack uses several different models, obtained from, e.g., Model Zoo \cite{modelzoo}, queries all models, and finally chooses the output that was given by most of the models. If one of the models is watermarked and triggered with a specific input for the watermark extraction process, most likely only the watermarked model will predict the chosen label, while the remaining models will predict the true label. Therefore, the trigger output will not be returned, and the verification fails.
The detection attack tries to avoid a trigger response; it trains a neural network,i.e., the \textit{detector}, which predicts if the query is intending to trigger a watermark. If the input is recognized as a trigger image, a different or no class at all can be returned. The detector is a binary classifier that needs to distinguish between \textit{clean-} and \textit{trigger} input. Clean input is collected from other public datasets. 
Trigger inputs are generated from a portion of these samples. It should be noted that this kind of attack is not able to invalidate \textit{pattern-based}, \textit{noise-based} and \textit{in-distribution} watermarks, as the detector cannot be trained well for watermark detection without further information about the watermark.

% ----- Namba: Robust and Undetectable White-box Watermarks for Deep Neural Networks
Namba et al. \cite{namba_robust_2019} proposed a watermark invalidation attack called \textit{query modification processing}, consisting of two steps: trigger sample detection and query modification via autoencoder (AE). An autoencoder can reduce the effect of trigger images by diluting the pattern embedded in the original image, or by eliminating the embedded noise. Because the application of an autoencoder to non-trigger images impacts the performance of the model on these images negatively, it is not recommended to use the AE on every query. Similar to \cite{hitaj_evasion_2019}, the authors propose to first detect if the input could be a trigger image queried during a watermark verification process. They suggest three ways to perform the detection: (i) measuring the effect of the autoencoder to the image in the input space, (ii) measuring the effect in the output space, or (iii) both.
The authors demonstrated to invalidate the watermarks created by \cite{zhang_protecting_2018,merrer_adversarial_2019,rouhani_deepsigns_2019}.

\subsection{Access Control Invalidation}
Besides breaking a (potentially insecure) mechanism underlying encryption or obfuscation, an obvious attack on an access control system is trying to guess a valid \textit{secret}.
The schemes presented in \Cref{sec:otherIPP} all demonstrate that using a wrong secret entails a large drop in fidelity, often to the level of a random classifier. Thus, their vulnerability depends on aspects like management of the secret or brute-force attacks trying random secrets. The success of these attacks depends on the size or complexity of the secret employed; thus, this can be a decisive factor in selecting a scheme.

Besides these, the most widely studied attack to render access control to ML models ineffective is fine-tuning.
Most proposed schemes test for this attack. Xu et al. \cite{xu_deepobfuscation_2018} showed that their scheme is to some extent resistant to fine-tuning, thus re-using the pirated network for other tasks is disadvantageous.
Chakraborty et al. \cite{chakraborty_hardware-assisted_2020}, however, show that a fine-tuning attack using 10\% of the dataset restores the accuracy to 4-11\% of the original accuracy. While this is still potentially large enough to bring little value, it also does not render the network completely unusable.
This is addressed by Alam et al. \cite{alam_nn-lock_2022}, who showed that a model fine-tuning attack -- with 10\% of the initial number of samples -- does not improve the random model accuracy of using a wrong secret.
Also, AprilPyone et al. \cite{aprilpyone_training_2020} demonstrated robustness to fine-tuning.
Chen et al.'s \cite{chen_protect_2018} scheme is vulnerable to a powerful attacker that can observe input-output patterns from the transformation module; depending on their amount, they can then restore prediction accuracy to within 5-15\% of the original one -- which might still be too large to make the attack not worthwhile.

\subsection{Other Attack Considerations}
Kupek et al. \cite{kupek_difficulty_2020} studied defenses against adversarial attacks. 
They investigate to what extent secret (defense) parameters -- which have an effect on the model parameters, e.g., a weights modification during fine-tuning with an additional loss function -- can be estimated by an attacker.
If this estimation succeeds, the attack can be tailored to better circumvent the defense.
While not primarily studied in the IPP context, this type of parameter estimation could be utilized in attacks against some of the schemes discussed in this paper, similar to the vulnerability mentioned by Wang et al. \cite{wang_riga_2021}.

It can be observed that in contrast to reactive methods like watermarking, there is, at the time of writing, too few works that evaluate pro-active schemes -- there is especially a lack of works that independently evaluate schemes, i.e. an evaluation done by others than the original authors of the scheme. This might be due to access control techniques being generally newer and, therefore, less explored. However, it indicates a need for a more systematic and thorough theoretical and empirical evaluation of the proposed schemes.

\section{Guidelines on Choosing an IPP method}
\label{sec:guidelines}

ML models are certainly an IP that needs to be secured when making it publicly available. Model owners that want to determine which security measures to take are confronted with a variety of possibilities, which we analyzed and systematized in this paper. Based on this work, we can derive a set of guiding questions that will help to decide which action to take:

\textit{Do I want to proactively protect my model from malicious users or react in case of a threat event?}
If proactively, one should consider model access mechanisms or unrobust models (cf. \cref{sec:otherIPP}); if reactively, Watermarking would be an appropriate choice (cf. \cref{sec:watermarking}).

\textit{When needing ownership verification, can I ensure to get full access to the adversary model?}
If access is not ensured, a black-box approach is the appropriate choice. Otherwise, both white- (cf. \cref{sec:whitebox}) and black-box (cf. \cref{sec:blackbox}) are suitable, where white-box watermarking schemes tend to have higher fidelity than black-box watermarking.

\textit{How am I going to distribute my model?}
In case of an API service, one should be aware of Model Stealing attacks (cf. \cref{sec:wm:countering_extraction}). Most of the introduced methods are not robust against this type of attack, except of EWE \cite{jia_entangled_2021} and DAWN \cite{szyller_dawn_2021}. When distributing the full model, one should choose an IPP that is robust against model modifications -- since that is what an attacker would most likely do before re-distributing -- or decide whether access control is of importance.

\textit{If distributed to multiple users, do I need to be able to trace back a malicious user?}
If yes, one should consider embedding fingerprints (user-specific watermarks) each time before distributing a model to the users. DeepMarks \cite{chen_deepmarks_2019} is so far the only explicitly designed fingerprinting framework that allows unique fingerprint embedding and also detects if users collaborated. However, we believe that other watermarking methods could be extended to fingerprinting, as fingerprinting is a user-specific watermark (cf. \cref{sec:fingerprinting:user-specific-wm}).

\textit{Is my model large enough to hold additional watermark- or model access information?}
Model owners should be aware that the larger a model, the better it will perform on fidelity, since the model has enough "space" for holding the additional information without compromising test accuracy.

\textit{Do I already have a trained model?}
Most watermarking methods and some model access techniques, e.g., \cite{tang_deep_2020, chen_protect_2018}, embed the information when training the model from scratch. Although it is possible to embed the information later on, fidelity is in this case often compromised. Some of the watermarking methods need an already trained model, e.g., \cite{merrer_adversarial_2019} for generating adversarial examples, or \cite{namba_robust_2019}. If in possession of an already trained model, the model owner can utilize those watermarking methods or implement every other watermarking method, but has to be aware of the fidelity loss.
Regarding model access techniques, similar observations hold true -- some methods, such as \cite{tang_deep_2020, chen_protect_2018}, adapt the training process with an additional regularizer to embed information, and thus most often lead to higher fidelity if they are already employed during training, and not during fine-tuning on a previously trained model.

\textit{How much effort can I expect an attacker to spend on defeating my security mechanism?}
Attacks require different amounts of training time, (substitute) training data, different levels of access, etc. (cf. \cref{sec:attacks}).
This needs to be balanced with the (expected) value which the attack might yield.
When choosing an appropriate watermarking method, one should be aware that most methods face a trade-off between robustness and fidelity. 

There are watermarking techniques which have not been broken so far, such as the schemes in \cite{yang_effectiveness_2019, li_piracy_2020, guo_evolutionary_2019, zhu_secure_2020, li_how_2019, chen_blackmarks_2019, zhong_protecting_2020, zhang_deeptrigger_2020, xu_identity_2020, wang_riga_2021, wang_watermarking_2020, feng_watermarking_2020, jia_entangled_2021, szyller_dawn_2021, wu_watermarking_2021, quan_watermarking_2020, yu_artificial_2021, chen_specmark_2020, guan_reversible_2020} (cf. \cref{tab:requirement}, and \cref{tab:attacks}). 
However, it does not follow that these methods are more robust than others, especially as many of these schemes are rather new -- and many schemes have only been tested against some attacks.

Regarding model access, we note that most schemes have not been vetted against attacks developed by researchers other than the original authors of the scheme, and more empirical evaluation is required.
Thus, at this point, it is not possible to accurately estimate the required effort of an attacker.

\section{Conclusions} \label{sec:conclusions}
Intellectual property protection (IPP) for machine learning (ML) assets is a very active research field, but still in its infancy. With a growing number of threats discovered, novel protection methods proposed and counter-attacks developed, the lack of a unified view on the vulnerabilities hinders comprehensive approaches.
In this paper, we performed a systematic review of the field and provide a comprehensive taxonomy of IPP methods for ML models. We further categorized attacks on IPP mechanisms and discussed which specific mechanisms are affected by the attacks. This provides IP holders with a holistic overview of appropriate mechanisms so that they may perform a detailed investigation for a concrete setting.

We note that there is a lack of methods that holistically address multiple threats and attack models. Combining, e.g., model access control systems with other proactive measures such as unrobust models and embedded watermarks, would provide protection against multiple types of attacks. However, there might also be effects that multiple IPP strategies interfere with each other, especially if they need to modify similar aspects of the ML model.
With this survey and systematization of knowledge, we provide a starting point in this direction and inform about the complexity of a successful IPP of ML models. Future research is mandated to strengthen and extend evaluation frameworks for IPP methods in ML. 

In order to make future work on IPP protection methods and respective attacks more comparable, it is important to establish a benchmark setting with well-defined tasks, evaluation metrics, and artifacts.
Evaluating protection and attack methods on a common set of architectures and datasets fosters direct comparison. Thus, re-using previously utilized datasets and architectures is highly encouraged; if this is not possible, newly created artifacts (e.g., not yet employed datasets, model architectures, trained models and similar artifacts) need to be made available to the research community in an easy and reliable manner, with enough details to understand and reuse them.
Also, reproducibility of the experiments is vital in order to ensure that others can compare novel work to previously published results and, especially if new artifacts are used, are able to employ existing methods on these artifacts.
Thus, a detailed documentation of the training process and the hyper-parameters used is required. 

\balance 

\bibliography{TNNLS}
\bibliographystyle{hieeetr}

\begin{IEEEbiography}[{\includegraphics[width=1in,height=1.25in,clip,keepaspectratio]{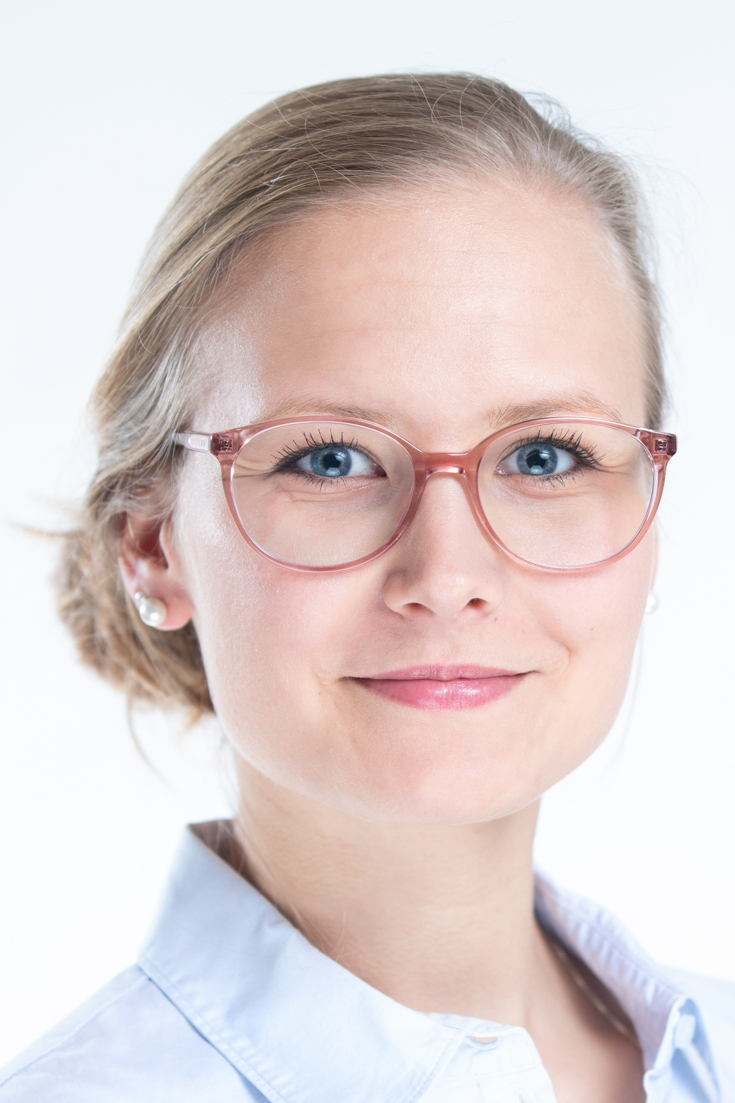}}]{Isabell Lederer} received her M.Sc. in technical mathematics from Technical University of Vienna in 2021. Her master thesis iss based on research on Intellectual Property Protection for Machine Learning models, in particular Watermarking methods for Convolutional Neural Networks. After her studies, she is pursuing a professional career as a Data Scientist.
\end{IEEEbiography}
%\vspace*{-1cm}

\begin{IEEEbiography}[{\includegraphics[width=1in,height=1.25in,clip,keepaspectratio]{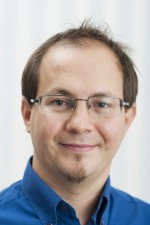}}]{Rudolf Mayer} is a senior researcher and lead of the machine learning and data management team at SBA Research, Vienna, Austria, and a lecturer at Vienna University of Technology. His research interests include information retrieval (focusing on text and music data), and machine learning. Specifically, he focuses on privacy-preserving data publishing and machine learning, as well as security aspects of machine learning (adversarial machine learning), and IP protection in machine learning processes.
\end{IEEEbiography}
%\vspace*{-1cm}

\begin{IEEEbiography}[{\includegraphics[width=1in,height=1.25in,clip,keepaspectratio]{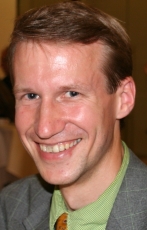}}]{Andreas Rauber} is professor at the Data Science Research Unit at the Department of Information Systems Engineering at the Vienna University of Technology; head of the Vienna Scientific Cluster Research Center; and key researcher at SBA Research. 
His research interests cover the broad scope of data science, ranging from reproducibility and transparency aspects in data analytics, and their realization in virtual research environments, to explainability and accountability in machine learning.
\end{IEEEbiography}
%\vspace*{-1cm}

\end{document}

%% file: fig-IPP-overview.tex
% ------ IPP overview figure
\begin{figure*}[t]

\begin{tikzpicture}
\node[inner sep=0pt] (overview) at (0,0) {\includegraphics[width=\textwidth]{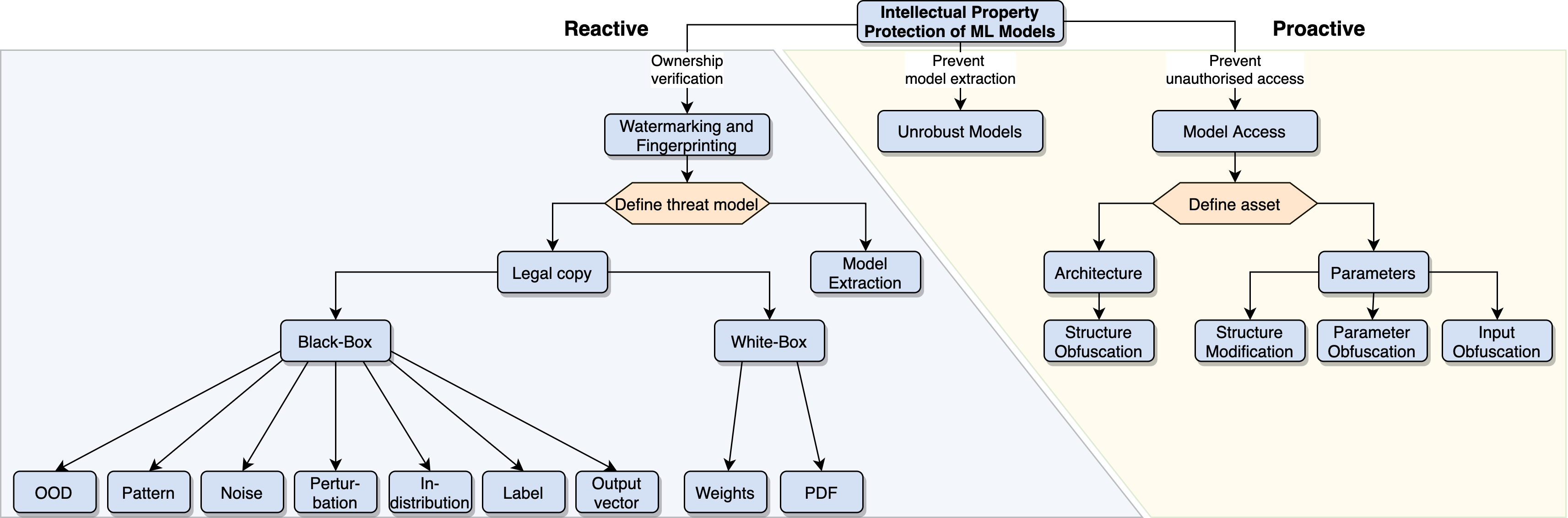}};
%OOD:
\node[inner sep=0pt] (cite1) at (-8.72,-3.1) {\tiny \cite{adi_turning_2018}};
\node[inner sep=0pt] (cite2) at (-8.42,-3.1) {\tiny \cite{zhang_protecting_2018}};
\node[inner sep=0pt] (cite22) at (-8.12,-3.1) {\tiny \cite{yang_effectiveness_2019}};
%pattern:
\node[inner sep=0pt] (cite3) at (-7.65,-3.1) {\tiny \cite{zhang_protecting_2018}};
\node[inner sep=0pt] (cite4) at (-7.35,-3.1) {\tiny \cite{li_piracy_2020}};
\node[inner sep=0pt] (cite5) at (-7.05,-3.1) {\tiny \cite{guo_watermarking_2018}};
\node[inner sep=0pt] (cite6) at (-7.5,-3.28) {\tiny \cite{guo_evolutionary_2019}};
\node[inner sep=0pt] (cite62) at (-7.2,-3.28) {\tiny \cite{tekgul_waffle_2021}};
%noise:
\node[inner sep=0pt] (cite7) at (-6.4,-3.1) {\tiny \cite{zhang_protecting_2018}};
\node[inner sep=0pt] (cite8) at (-6.1,-3.1) {\tiny \cite{zhu_secure_2020}};
%perturbation:
\node[inner sep=0pt] (cite9) at (-5.35,-3.1) {\tiny \cite{merrer_adversarial_2019}};
\node[inner sep=0pt] (cite10) at (-5.05,-3.1) {\tiny \cite{li_how_2019}};
\node[inner sep=0pt] (cite12) at (-5.29,-3.28) {\tiny FP: \cite{zhao_afa_2019}};
\node[inner sep=0pt] (cite13) at (-4.85,-3.28) {\tiny \cite{lukas_deep_2021}};
%in-distr.:
\node[inner sep=0pt] (cite14) at (-4.08,-3.1) {\tiny \cite{namba_robust_2019}};
%label:
\node[inner sep=0pt] (cite15) at (-3.15,-3.1) {\tiny \cite{zhong_protecting_2020}};
\node[inner sep=0pt] (cite16) at (-2.85,-3.1) {\tiny \cite{zhang_deeptrigger_2020}};

%output vector:
\node[inner sep=0pt] (cite162) at (-2.2,-3.1) {\tiny \cite{chen_blackmarks_2019}};
\node[inner sep=0pt] (cite161) at (-1.9,-3.1) {\tiny \cite{xu_identity_2020}};
\node[inner sep=0pt] (cite162) at (-1.6,-3.1) {\tiny \cite{yu_artificial_2021}};

%into weights:
\node[inner sep=0pt] (cite17) at (-0.8,-3.1) {\tiny \cite{uchida_embedding_2017}};
\node[inner sep=0pt] (cite18) at (-0.5,-3.1) {\tiny \cite{wang_riga_2021}};
\node[inner sep=0pt] (cite19) at (-0.8,-3.28) {\tiny \cite{wang_watermarking_2020}};
\node[inner sep=0pt] (cite20) at (-0.5,-3.28) {\tiny \cite{feng_watermarking_2020}};
%into pdf:
\node[inner sep=0pt] (cite21) at (0.45,-3.1) {\tiny \cite{rouhani_deepsigns_2019}};
\node[inner sep=0pt] (cite22) at (0.3,-3.28) {\tiny FP: \cite{chen_deepmarks_2019}};
%Model Extraction:
\node[inner sep=0pt] (cite23) at (0.8,-0.54) {\tiny \cite{jia_entangled_2021}}; 
\node[inner sep=0pt] (cite24) at (1.1,-0.54) {\tiny \cite{szyller_dawn_2021}};
% man könnte noch diese hinzufügen, aber lasse ich für die version erstmal...
%\node[inner sep=0pt] (cite241) at (1.6,-1.4) {\tiny \cite{wu_watermarking_2021}};
%\node[inner sep=0pt] (cite242) at (1.9,-1.4) {\tiny \cite{zhang_model_2020}};

% Unrobust models
 \node[inner sep=0pt] (cite25) at (2.05,1.08) {\tiny \cite{szentannai_preventing_2020}};
 
%Structure Obfuscation:
\node[inner sep=0pt] (cite101) at (3.65,-1.33) {\tiny \cite{xu_deepobfuscation_2018}};
 
% Structure Modification:
\node[inner sep=0pt] (cite102) at (5.1,-1.33) {\tiny \cite{fan_rethinking_2019}};
\node[inner sep=0pt] (cite1022) at (5.4,-1.33) {\tiny \cite{lim_protect_2022}};
\node[inner sep=0pt] (cite1023) at (5.7,-1.33) {\tiny \cite{sun_convolutional_2018}};

% Parameter Encryption & Obfuscation:
\node[inner sep=0pt] (cite103) at (6.2,-1.33) {\tiny \cite{gomez_security_2019}};
\node[inner sep=0pt] (cite104) at (6.5,-1.33) {\tiny \cite{chakraborty_hardware-assisted_2020}};
\node[inner sep=0pt] (cite105) at (6.8,-1.33) {\tiny \cite{alam_nn-lock_2022}};
\node[inner sep=0pt] (cite106) at (7.1,-1.33) {\tiny \cite{tang_deep_2020}};
\node[inner sep=0pt] (cite107) at (7.4,-1.33) {\tiny \cite{lin_chaotic_2020}};

%Input Obfuscation:
\node[inner sep=0pt] (cite108) at (8.1,-1.33) {\tiny \cite{aprilpyone_training_2020}};
\node[inner sep=0pt] (cite109) at (8.4,-1.33) {\tiny \cite{chen_protect_2018}};

\end{tikzpicture}

%    \caption{Taxonomy of IPP protection mechanisms for ML models.}
    \caption{Taxonomy of Intellectual Property Protection mechanisms for Machine Learning models.}
        \label{fig:overview}
\end{figure*}
% ------ end
% Ucomment for local usage
% \input{bibliography.bbl}

%% file: fig-InformationHiding_ML_Process_wide.tex
% ------ ML process overview figure

\begin{tikzpicture}
\node[inner sep=0pt] (mlProcess) at (-2.1,0) {\includegraphics[width=.7\linewidth]{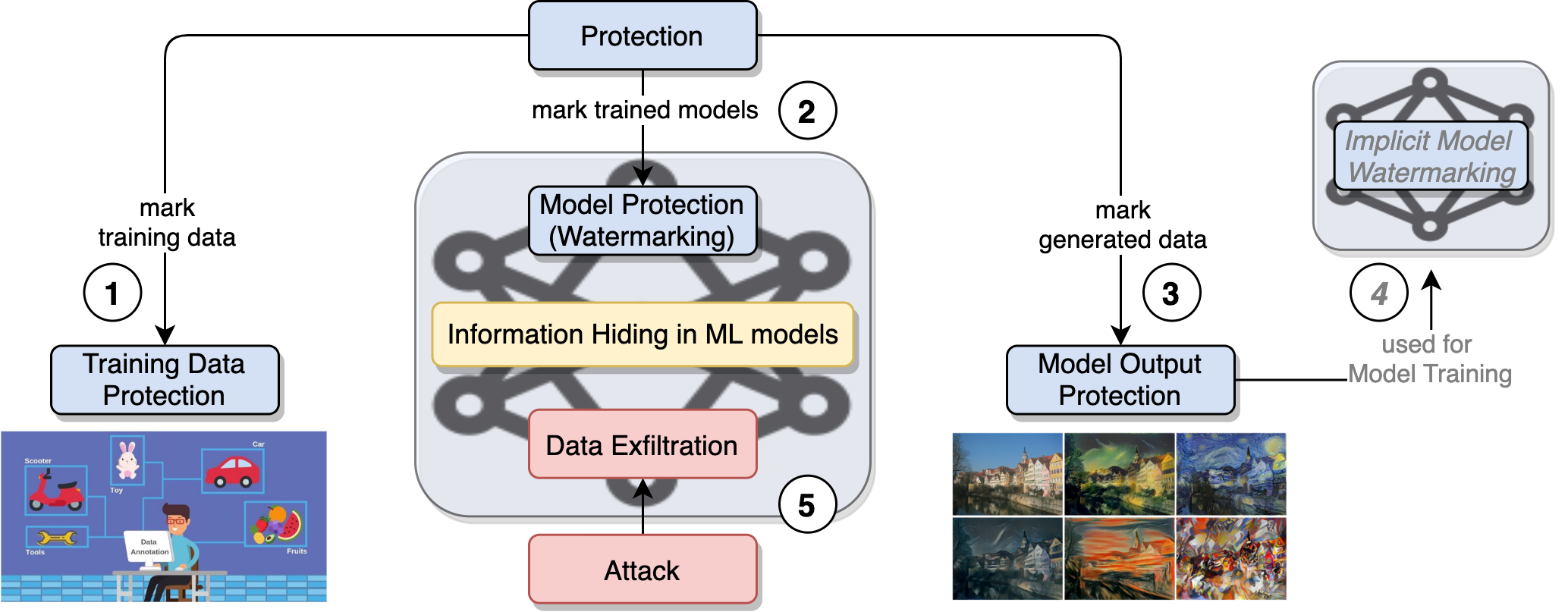}};

%MarkTrainingData:
\node[inner sep=0pt] (cite200) at (-6.8,0.07) {\tiny \cite{sablayrolles_radioactive_2020}};
%MarkOutputData:
% marks output, doesn't talk about model training
\node[inner sep=0pt] (cite211) at (1.5,0.07) {\tiny \cite{abdelnabi_adversarial_2021}};

%specifically for model training!
\node[inner sep=0pt] (cite212) at (3.5,0.2) {\tiny \cite{szyller_dawn_2021}};

% marks all output for surrogate model training
\node[inner sep=0pt] (cite213) at (3.5,0.0) {\tiny \cite{zhang_model_2020}};
\node[inner sep=0pt] (cite214) at (3.5,-0.2) {\tiny \cite{wu_watermarking_2021}};
%OtherInformationHiding
\node[inner sep=0pt] (cite225) at (-2.5,-1.5) {\tiny \cite{song_machine_2017}};
\end{tikzpicture}

%% file: tab-requirements.tex
\begin{table*}[t]
\centering
\caption{Requirements for Watermarking techniques. The notation is not consistent throughout the papers, but the terms in the left column are the most prominent ones. These requirements mostly apply also to Fingerprinting methods}

\rowcolors{2}{gray!15}{white}

\begin{tabular}{|p{1.35cm}|p{8.9cm}|p{6.95cm}|}

\rowcolor{gray!15}
\hline
\textbf{Property}               & \textbf{Description}                                                                                              & \textbf{Other terms used in papers}                        \\ \hline
Effectiveness                   & The model owner should be able to prove ownership anytime and multiple times if needed                            & Authentication \cite{li_piracy_2020}, Functionality \cite{li_how_2019}                              \\ \hline
Fidelity                        & The accuracy of the model should not be degraded after embedding the watermark                                    & Funcionality-preserving \cite{li_piracy_2020, wang_riga_2021, adi_turning_2018}, Loyalty \cite{merrer_adversarial_2019}, Utility \cite{szyller_dawn_2021}
\scriptsize{(Image WM: Transparency \cite{potdar_survey_2005}; relational data: Usability \cite{kamran_comprehensive_2018})}                \\ \hline
Robustness                      & The embedded watermark should resist a designated class of transformations                                        & Unremovability \cite{adi_turning_2018, szyller_dawn_2021}                                            \\ \hline
Security                        & The watermark should be secure against brute-force or specifically crafted evasion attacks                        & Secrecy \cite{yu_artificial_2021}, Unforgeability \cite{adi_turning_2018, wang_riga_2021}                                    \\ \hline
Legality                        & An adversary cannot produce a watermark for a model that was already watermarked by the model owner               & Ownership piracy resilient \cite{adi_turning_2018, wang_riga_2021}, Non-ownership piracy \cite{szyller_dawn_2021}           \\ \hline
Integrity                       & The watermark verification process should have a negligible false positive rate                                   & Low false positive rate \cite{guo_evolutionary_2019, guo_watermarking_2018}, Non-trivial ownership \cite{li_piracy_2020, adi_turning_2018, wang_riga_2021}, Uniqueness \cite{quan_watermarking_2020} \\ \hline
Reliability                     & The watermark verification process should have a negligible false negative rate                                   & Credibility \cite{chen_blackmarks_2019}                                                \\ \hline
Efficiency                      & The watermarking embedding and verification process should be fast                                                &                                                            \\ \hline
Capacity                        & The watermarking scheme should be capable of embedding a large amount of information                              & Payload \cite{guo_watermarking_2018}                                                    \\ \hline

\end{tabular}
\label{tab:requirement}
\end{table*}

%% file: tab-summary-req.tex
\begin{table*}[t]
\centering
% No punctuation in last sentence + caption should be set as an inverted pyramid. see IEEE editorial style manual: http://journals.ieeeauthorcenter.ieee.org/wp-content/uploads/sites/7/IEEE-Editorial-Style-Manual_081920.pdf
\caption{Requirements met by watermarking and fingerprinting schemes.  We distinguish two degrees:
%$\sim$ indicates a claim -- the authors of the scheme claim that the scheme fulfills this property; \checkmark indicates a demonstration -- the authors show empirically that the property is fulfilled
$\sim$ indicates: the respective authors claim the scheme fulfills this property; \checkmark indicates: the authors show empirically to which extent the property is fulfilled
}

\rowcolors{2}{white}{gray!15}

\setlength\tabcolsep{3pt}

\begin{tabular}{|l|c|c|c|c|c|c|c|c|c|c|c|c|c|c|c|c|c|c|c|c|c|c|c|c|c|c|c|c|c|c|c|}
\rowcolor{gray!15}
\hline                                     & \multicolumn{7}{l|}{\textbf{white-box}}                                                                             & \multicolumn{24}{l|}{\textbf{black-box}}                                                                                                                                                                                                                                                                                                                                                                  \\ \cline{2-32} 
%                                     & \multicolumn{2}{l|}{\textbf{into pdf}} & \multicolumn{4}{l|}{\textbf{into weights}}                   & \textbf{?} & \textbf{OOD, pattern, noise} & \textbf{OOD} & \textbf{pattern} & \textbf{}         &                   &        & \multicolumn{5}{l|}{\textbf{perturbation}}           & \textbf{in-distr} & \multicolumn{2}{l|}{\textbf{against ME}} & \multicolumn{3}{l|}{\textbf{labels}} & \multicolumn{4}{l|}{\textbf{extending existing work}}          & \multicolumn{3}{l|}{\textbf{image processing}}            \\ \hline
%
\hline
\multicolumn{1}{|l|}{\textbf{Property}}      & {\tiny\cite{rouhani_deepsigns_2019}}         & {\tiny\cite{chen_deepmarks_2019}}      & {\tiny\cite{uchida_embedding_2017}}     & {\tiny\cite{wang_riga_2021}} & {\tiny\cite{wang_watermarking_2020}} & {\tiny\cite{feng_watermarking_2020}}       & {\tiny\cite{chen_specmark_2020}}   & {\tiny\cite{zhang_protecting_2018}}                        & {\tiny\cite{adi_turning_2018}}          & {\tiny\cite{li_piracy_2020}}       & {\tiny\cite{guo_watermarking_2018}} & {\tiny\cite{guo_evolutionary_2019}} & {\tiny\cite{zhu_secure_2020}}    & {\tiny\cite{merrer_adversarial_2019}} & {\tiny\cite{li_how_2019}} & {\tiny\cite{chen_blackmarks_2019}}  & {\tiny\cite{zhao_afa_2019}} & {\tiny\cite{lukas_deep_2021}} & {\tiny\cite{cao_ipguard_2021}} & {\tiny\cite{namba_robust_2019}}             & {\tiny\cite{szyller_dawn_2021}}       & {\tiny\cite{jia_entangled_2021}}                     & {\tiny\cite{zhong_protecting_2020}}     & {\tiny\cite{zhang_deeptrigger_2020}}    & {\tiny\cite{xu_identity_2020}}      & {\tiny\cite{yang_effectiveness_2019}}      & {\tiny\cite{guan_reversible_2020}} & {\tiny\cite{yu_artificial_2021}}  & {\tiny\cite{wu_watermarking_2021}} &   {\tiny\cite{zhang_model_2020}}     &  {\tiny\cite{quan_watermarking_2020}}       \\ \hline

\multicolumn{1}{|l|}{Effectiveness} & \checkmark      & \checkmark                 & \checkmark & \checkmark   &  \checkmark                   & \checkmark & \checkmark           & \checkmark                   & \checkmark   & \checkmark             & \checkmark              & \checkmark              & \checkmark   & \checkmark   & \checkmark    &   \checkmark          & \checkmark      & \checkmark & \checkmark   & \checkmark              & \checkmark          & \checkmark                    & \checkmark      & \checkmark           & \checkmark    & \checkmark      & \checkmark             & \checkmark                & \checkmark  &  \checkmark  & \checkmark                 \\ \hline

\multicolumn{1}{|l|}{Fidelity}      & \checkmark            & \checkmark                 & \checkmark       & \checkmark         & \checkmark                & \checkmark       & \checkmark       & \checkmark                         & \checkmark         & \checkmark             & \checkmark              & \checkmark              & \checkmark   & \checkmark   & \checkmark    & \checkmark        & \checkmark   & \checkmark & \checkmark & \checkmark              & \checkmark          & \checkmark                    & \checkmark      & \checkmark           & \checkmark    & \checkmark      & \checkmark             & \checkmark                     & \checkmark                   &        \tiny N/A             & \checkmark       \\ \hline

% \multicolumn{1}{|l|}{Robustness}    & FT, MC, OW      & FT, MC, OW           & FT, MC     & FT, MC, OW   & FT, MC              & FT, MC, OW & FT, MC, TL & FT, MC                       & FT, TL       & FT, MC           & $\sim$             & FT                & FA, TL & FT, MC & FT      & FT, MC , OW & FT, MC    & \checkmark    & FT, MC            & ME            & ME, MC, FT, NC, FP       & FT, MC    & FT, MC, OW     & MC      & FT, MC, D &                  & FT,  MC          & FT, MC       & cropping, adding noise & \checkmark                & FT, MC, OW \\ \hline

\multicolumn{1}{|l|}{Robustness}    & \checkmark & \checkmark & \checkmark & \checkmark & \checkmark & \checkmark & \checkmark & \checkmark & \checkmark & \checkmark & $\sim$ & \checkmark & \checkmark & \checkmark & \checkmark     & \checkmark & \checkmark & \checkmark & \checkmark   & \checkmark  & \checkmark  & \checkmark  & \checkmark & \checkmark  & \checkmark & \checkmark &                  & \checkmark  & \checkmark & \checkmark & \checkmark \\ \hline

\multicolumn{1}{|l|}{Security}      & \checkmark            & \checkmark                 & $\sim$      & $\sim$        & $\sim$               &            & $\sim$      & \checkmark                         & $\sim$        &                  & \checkmark              &                   & \checkmark   & $\sim$  & \checkmark    & \checkmark        &           &    &     &                   &               &                         &           & \checkmark           &         &           &                  & \checkmark                     & \checkmark                   &                     &            \\ \hline

\multicolumn{1}{|l|}{Legality}      &                 &                      &            & $\sim$        &                     &            &            &                              &              &                  &                   &                   &        &        & \checkmark    &             &           &     &    &                   & $\sim$         &                         &           &                &         &           &                  &                      &                        &                     &            \\ \hline

\multicolumn{1}{|l|}{Integrity}     & \checkmark            & \checkmark                 &            & \checkmark         &                     &            & \checkmark       &                              & \checkmark         & \checkmark             & \checkmark              & \checkmark              &        &        & \checkmark    & \checkmark        & \checkmark      & \checkmark  &   &                   &               &                         &           & \checkmark           &         &           &                  &                            &                        &                     &            \\ \hline

\multicolumn{1}{|l|}{Reliability}   & \checkmark            & \checkmark                 &            &              &                     &            & $\sim$      &                              &              &                  &                   &                   &        &        &         & \checkmark        & \checkmark      & \checkmark  &   &                   & \checkmark          &                         &           &                &         &           &                  &                      &                        &                     &            \\ \hline

\multicolumn{1}{|l|}{Efficiency}    & \checkmark            & \checkmark                 & $\sim$      &              & $\sim$               & \checkmark       & \checkmark       &                              &              &                  &                   &                   &        & $\sim$  &         & \checkmark        &           &    &  \checkmark   &                   & \checkmark          &                         & \checkmark      &                &         &           &                  &                       &                        &                     &            \\ \hline

\multicolumn{1}{|l|}{Capacity}      & \checkmark            &                      & \checkmark       &              & \checkmark                & \checkmark       &            &                              &              &                  & \checkmark              &                   &        &        &         & \checkmark        &           &     &     &                   &               &                         &           &                &         &           &                  & \checkmark                  & \checkmark                   &                     & $\sim$      \\ \hline
\end{tabular}

\label{tab:summary-req}
\end{table*}

%% file: tab-summary-req-accessControl.tex
\begin{table*}[t]
\centering
\caption{Requirements met by access control schemes. We distinguish two degrees:
$\sim$ indicates: the respective authors claim the scheme fulfills this property; \checkmark indicates: the authors show empirically to which extent the property is fulfilled}
\label{tab:summary-req-accessControl}

\rowcolors{2}{gray!15}{white}

\setlength\tabcolsep{3pt}

\begin{tabular}{|l|c|ccc|ccccc|cc|}
\rowcolor{gray!15}
\hline
 &
  \textbf{\begin{tabular}[c]{@{}c@{}}Structure\\ Obfuscation\end{tabular}} &
  \multicolumn{3}{c|}{\textbf{\begin{tabular}[c]{@{}c@{}}Structure \\ Modification\end{tabular}}} &
  \multicolumn{5}{c|}{\textbf{\begin{tabular}[c]{@{}c@{}}Parameter Encryption\\ \& Obfuscation\end{tabular}}} &
  \multicolumn{2}{c|}{\textbf{\begin{tabular}[c]{@{}c@{}}Input\\ Obfuscation\end{tabular}}} \\ \hline
\textbf{Property} &
  {\tiny\cite{xu_deepobfuscation_2018}} &
  \multicolumn{1}{c|}{{\tiny\cite{fan_rethinking_2019}}} &
  \multicolumn{1}{c|}{{\tiny\cite{lim_protect_2022}}} &
  {\tiny\cite{sun_convolutional_2018}} &
  \multicolumn{1}{c|}{{\tiny\cite{gomez_security_2019}}} &
  \multicolumn{1}{c|}{{\tiny\cite{chakraborty_hardware-assisted_2020}}} &
  \multicolumn{1}{c|}{{\tiny\cite{alam_nn-lock_2022}}} &
  \multicolumn{1}{c|}{{\tiny\cite{tang_deep_2020}}} &
  {\tiny\cite{lin_chaotic_2020}} &
  \multicolumn{1}{c|}{{\tiny\cite{aprilpyone_training_2020}}} &
  {\tiny\cite{chen_protect_2018}} \\ \hline
Fidelity &
  \checkmark &
  \multicolumn{1}{c|}{\checkmark} &
  \multicolumn{1}{c|}{\checkmark} &
  \checkmark &
  \multicolumn{1}{c|}{\checkmark} &
  \multicolumn{1}{c|}{\checkmark} &
  \multicolumn{1}{c|}{\checkmark} &
  \multicolumn{1}{c|}{\checkmark} &
  $\sim$ &
  \multicolumn{1}{c|}{\checkmark} &
  \checkmark \\ \hline
Protection Effectiveness &
  $\sim$ &
  \multicolumn{1}{c|}{\checkmark} &
  \multicolumn{1}{c|}{\checkmark} &
  \checkmark &
  \multicolumn{1}{c|}{} &
  \multicolumn{1}{c|}{\checkmark} &
  \multicolumn{1}{c|}{\checkmark} &
  \multicolumn{1}{c|}{\checkmark} &
  \checkmark &
  \multicolumn{1}{c|}{\checkmark} &
  \checkmark \\ \hline
Robustness &
   &
  \multicolumn{1}{c|}{\checkmark} &
  \multicolumn{1}{c|}{\checkmark} &
   &
  \multicolumn{1}{c|}{} &
  \multicolumn{1}{c|}{} &
  \multicolumn{1}{c|}{\checkmark} &
  \multicolumn{1}{c|}{\checkmark} &
   &
  \multicolumn{1}{c|}{} &
  \checkmark \\ \hline
Efficiency &
  \checkmark &
  \multicolumn{1}{c|}{} &
  \multicolumn{1}{c|}{} &
   &
  \multicolumn{1}{c|}{} &
  \multicolumn{1}{c|}{\checkmark} &
  \multicolumn{1}{c|}{\checkmark} &
  \multicolumn{1}{c|}{} &
  \checkmark &
  \multicolumn{1}{c|}{} &
   \\ \hline
\end{tabular}
\end{table*}

%% file: tab-attacks-transpose-new.tex
\begin{table*}[t]
\caption{Which attack defeats which watermarking technique based on the evaluation of the papers. A $\sim$ denotes that the authors claim that their attack can be extended easily to defeat this watermarking technique but did not provide an evaluation for that.} 

\centering
\resizebox{\textwidth}{!}{\begin{tabular}{llll|l|l|l|l|l|l|}
\cline{5-10}
                                                              &                                                    &                             &  & \multicolumn{6}{l|}{Watermarking techniques}                                                                          \\ \cline{2-10} 
    \multicolumn{1}{l|}{}                                                          & \multicolumn{1}{l|}{Attack goal}                                                   & \multicolumn{1}{l|}{Attack technique}                             & Attack paper            & OOD & pattern    & noise & perturb. & in-distr. & regulariser          \\ \hline
\multicolumn{1}{|l|}{\multirow{12}{*}{\rotatebox[origin=c]{90}{Attacks on watermarks}}} & \multicolumn{1}{l|}{\multirow{2}{*}{Invalidation}} & \multicolumn{1}{l|}{Substitute ensemble \& detector}  & Hitaj et al. \cite{hitaj_evasion_2019}        & \cite{adi_turning_2018}, \cite{zhang_protecting_2018}$\sim$        &            &       & \cite{merrer_adversarial_2019}$\sim$       &                 &                            \\ \cline{3-10} 
\multicolumn{1}{|l|}{}                                        & \multicolumn{1}{l|}{}                              & \multicolumn{1}{l|}{Query pre-processing}           & Namba et al. \cite{namba_robust_2019}        & \cite{zhang_protecting_2018}               & \cite{zhang_protecting_2018}      & \cite{zhang_protecting_2018} & \cite{merrer_adversarial_2019}       &                 & \cite{rouhani_deepsigns_2019}                    \\ \cline{2-10} 
\multicolumn{1}{|l|}{}                                        & \multicolumn{1}{l|}{Overwriting}                   &  \multicolumn{1}{l|}{Embed new watermark}                            & Li et al. \cite{li_piracy_2020}           & \cite{adi_turning_2018}, \cite{zhang_protecting_2018}          & \cite{zhang_protecting_2018}      & \cite{zhang_protecting_2018} &              &                 &                            \\ \cline{2-10} 
\multicolumn{1}{|l|}{}                                        & \multicolumn{1}{l|}{\multirow{2}{*}{Detection}}                     & \multicolumn{1}{l|}{\multirow{2}{*}{Property inference}}     & Shafieinejad et al. \cite{shafieinejad_robustness_2021} & \cite{adi_turning_2018}, \cite{zhang_protecting_2018}          & \cite{guo_watermarking_2018}, \cite{zhang_protecting_2018} & \cite{zhang_protecting_2018} &              &                 &                            \\ \cline{4-10} 
\multicolumn{1}{|l|}{}                                        & \multicolumn{1}{l|}{}                              &  \multicolumn{1}{l|}{}                            & Wang et al. \cite{wang_riga_2021}  &                     &            &       &              &                 & \cite{uchida_embedding_2017}, \cite{rouhani_deepsigns_2019}            \\ \cline{2-10} 
\multicolumn{1}{|l|}{}                                        & \multicolumn{1}{l|}{Detection, removal}            & \multicolumn{1}{l|}{Analyse weights variance}     & Wang et al. \cite{wang_attacks_2019} &                     &            &       &              &                 & \cite{uchida_embedding_2017}                     \\ \cline{2-10} 
\multicolumn{1}{|l|}{}                                        & \multicolumn{1}{l|}{\multirow{4}{*}{Removal}}      & \multicolumn{1}{l|}{\multirow{4}{*}{Fine-Tuning}} & Liu et al. \cite{liu_removing_2021}          & \cite{adi_turning_2018}, \cite{zhang_protecting_2018}          & \cite{guo_watermarking_2018}, \cite{zhang_protecting_2018} & \cite{zhang_protecting_2018} &              &                 &                            \\ \cline{4-10} 
\multicolumn{1}{|l|}{}                                        & \multicolumn{1}{l|}{}                              & \multicolumn{1}{l|}{}                             & Aiken et al. \cite{aiken_neural_2021}        & \cite{adi_turning_2018}, \cite{zhang_protecting_2018}          & \cite{zhang_protecting_2018}      & \cite{zhang_protecting_2018} &              &                 &                            \\ \cline{4-10} 
\multicolumn{1}{|l|}{}                                        & \multicolumn{1}{l|}{}                              &  \multicolumn{1}{l|}{}                            & Guo et al. \cite{guo_fine-tuning_2021}          & \cite{adi_turning_2018}                 & \cite{zhang_protecting_2018}      &       & \cite{merrer_adversarial_2019}       &                 &                            \\ \cline{4-10} 
\multicolumn{1}{|l|}{}                                        & \multicolumn{1}{l|}{}                              &  \multicolumn{1}{l|}{}                            & Chen et al. \cite{chen_refit_2021}        & \cite{adi_turning_2018}, \cite{zhang_protecting_2018}          & \cite{zhang_protecting_2018}      &       & \cite{merrer_adversarial_2019}       & \cite{namba_robust_2019}           &                            \\ \cline{3-10} 
\multicolumn{1}{|l|}{}                                        & \multicolumn{1}{l|}{}                      & \multicolumn{1}{l|}{Distillation}          & Yang et al. \cite{yang_effectiveness_2019}         &                     &            &       &              &                 & \cite{uchida_embedding_2017}, \cite{rouhani_deepsigns_2019}, \cite{chen_deepmarks_2019} \\ \hline
\end{tabular}
}

\label{tab:attacks}
\end{table*}

% \cite{hitaj_evasion_2019}
% \cite{namba_robust_2019}
% \cite{li_piracy_2020}
% \cite{sakazawa_visual_2019}
% \cite{wang_attacks_2019}
% \cite{wang_riga_2021}
% \cite{shafieinejad_robustness_2021}
% \cite{liu_removing_2021}
% \cite{aiken_neural_2021}
% \cite{guo_fine-tuning_2021}
% \cite{chen_refit_2021}
% \cite{yang_effectiveness_2019}

% \cite{\cite{adi_turning_2018}_turning_2018}
% \cite{\cite{zhang_protecting_2018}_protecting_2018}
% \cite{merrer_adversarial_2019}
% \cite{guo_watermarking_2018}
% \cite{namba_robust_2019}
% \cite{\cite{rouhani_deepsigns_2019}_deepsigns_2019}
% \cite{\cite{uchida_embedding_2017}_embedding_2017}
% \cite{chen_deepmarks_2019}